\documentclass[11pt]{article}

\usepackage[preprint]{acl}

\usepackage{times}
\usepackage{latexsym}
\usepackage{amssymb}

\usepackage[T1]{fontenc}

\usepackage[utf8]{inputenc}

\usepackage{microtype}

\usepackage{inconsolata}

\usepackage{graphicx}

\usepackage{hyperref}
\usepackage{url}
\usepackage{booktabs}

\usepackage{amsmath}
\usepackage[most]{tcolorbox}
\newtcolorbox{methodbox}{
  colback=gray!8,
  colframe=gray!60,
  boxrule=0.6pt,
  arc=2pt,
  left=6pt,
  right=6pt,
  top=6pt,
  bottom=6pt
}

\usepackage{multirow}
\usepackage[table]{xcolor}
\usepackage{colortbl}
\usepackage{wrapfig}
\usepackage{tabularx}
\usepackage{array}
\usepackage{makecell}
\usepackage{pifont}
\usepackage{longtable}

\definecolor{lightgrayrow}{RGB}{245,245,245}
\definecolor{lightbluerow}{RGB}{245,248,255}

\title{Learning from Synthetic Data without Model Collapse in Iterative Instruction Tuning}

\author{
  \textbf{Xiaonan Luo\textsuperscript{1}},
  \textbf{Yue Huang\textsuperscript{1}},
  \textbf{Kehan Guo\textsuperscript{1}},
  \textbf{Ping He\textsuperscript{2}},
  \textbf{Chuan Zou\textsuperscript{3}},
  \textbf{Ting Hua\textsuperscript{1}},
  \textbf{Xiangliang Zhang\textsuperscript{1}\thanks{Corresponding author.}}
\\
\\
  \textsuperscript{1}University of Notre Dame,
  \textsuperscript{2}Vanderbilt University,
  \textsuperscript{3}University of Pennsylvania
}

\begin{document}
\maketitle
\begin{abstract}
Model collapse is a central challenge in learning from synthetic data: as later-generation large language models (LLMs) are trained on an increasing proportion of model-generated data, performance can degrade due to narrowed coverage and accumulated bias. Existing work mainly studies how to bound this degradation. In iterative model evolution, however, the more meaningful objective is to ensure that each successive model improves over its predecessor, which requires diagnosing collapse at a granularity that is actionable for data curation. We study this problem in synthetic data self-improving for instruction tuning. We show that collapse in this setting is not simply uniform performance degradation, but can appear as a polarization of competence, where synthetic training reinforces already strong skills while further degrading weak ones. Motivated by this observation, we propose \textbf{KITE} (\textbf{K}nowledge-boundary \textbf{I}nstruction \textbf{T}uning via \textbf{E}xploration), a two-stage framework that combines failure-guided data generation with boundary-aware uncertainty curation. Experiments across several datasets and multiple open-source LLMs show that KITE yields more stable improvement than strong synthetic-data baselines.
\end{abstract}

\section{Introduction}

Synthetic data has become an increasingly important component in the post-training of large language models (LLMs)~\citep{liu2024best, schick-schutze-2021-generating, riabi-etal-2021-synthetic, zhao2024wildchat, wei2024simplesyntheticdatareduces, huang2025chemorch, huang2025datagen, lee2024aligning}. However, the growing dependence on synthetic data introduces a fundamental challenge: \emph{model collapse}~\citep{shumailov2024curserecursiontraininggenerated, dohmatob2024a, gerstgrasser2024is, zhu2025how, feng2025beyond, drayson-etal-2025-machine, amin2025escaping, kazdan2025collapse, dohmatob2025strong}. As synthetic data becomes an increasingly large component of future training corpora, new generations of LLMs will be optimized on data that partially reflects the outputs of previous generations. This can distort the effective training distribution, gradually reducing its diversity and fidelity to real-world data~\citep{shumailov2024curserecursiontraininggenerated}. Rather than preserving the full richness of human-produced corpora, the model may place increasing emphasis on recurring, high-probability patterns inherited from earlier synthetic generations. Such drift can accumulate across iterative training, ultimately reducing coverage of long-tail capabilities and harming performance on held-out tasks.

\begin{figure}[t]
    \centering
    \vspace{-8pt}
    \includegraphics[width=0.46\textwidth]{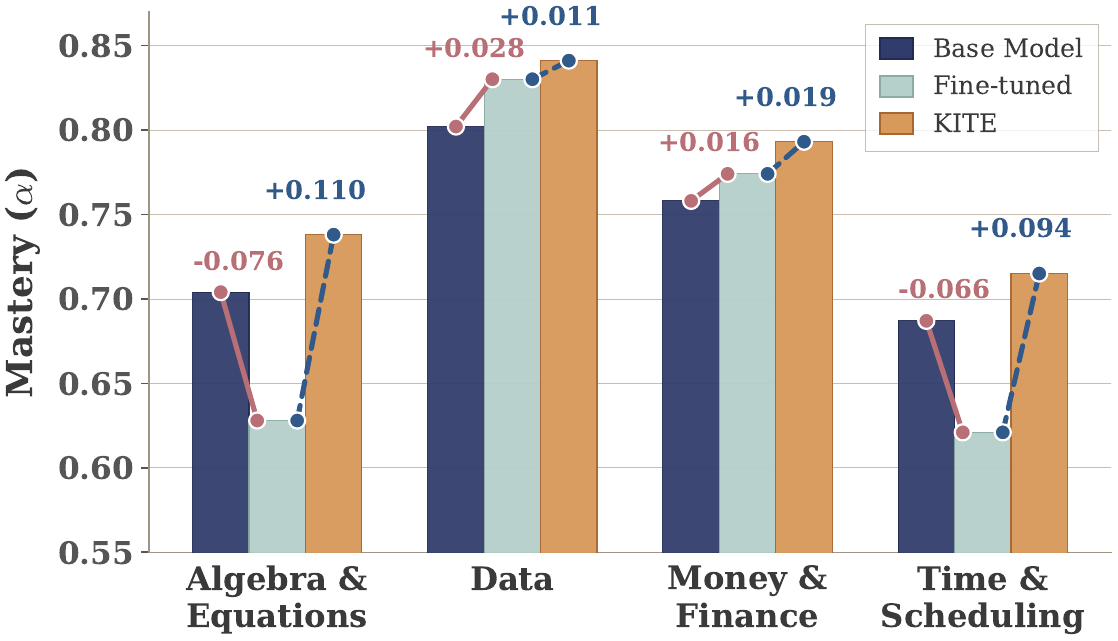}
    \vspace{-8pt}
    \caption{Selective skill-mastery profiles of Llama-3-8B-Instruct on GSM8K, comparing the base model with the model fine-tuned on accumulated synthetic data.}
    \label{fig:radar_skill}
    \vspace{-10pt}
\end{figure}

Recent studies advance the understanding of model collapse under synthetic data. \citet{shumailov2024curserecursiontraininggenerated} show that recursive training on model-generated data can induce a degenerative feedback loop, while \citet{dohmatob2024a} connect this phenomenon to changes in scaling laws driven by the loss of long-tail support. \citet{gerstgrasser2024is} further show that retaining real data alongside accumulated synthetic data can keep test error bounded, and \citet{zhu2025how} propose token-level editing as a practical way to synthesize text data while reducing collapse. Despite these advances, important gaps remain. \textbf{First}, existing work mainly asks how to control collapse so that degradation does not become unbounded. In iterative model evolution, however, the more meaningful objective is to ensure that each successive model improves over the previous generation. \textbf{Second}, existing analyses usually characterize collapse at an aggregate level, such as overall accuracy drop or long-tail narrowing, which is too coarse to guide synthetic data construction. For example, collapse may arise either because the data becomes too narrow and repetitive, which calls for data diversification, or because the model has specific weaknesses on a subset of skills, which calls for more targeted data construction. This motivates the need to \emph{diagnose collapse at a granularity that is actionable for data curation}. We study model collapse in the self-improving regime, where the \emph{instructions} that shape the training distribution are drawn from the model's own generations. Following common practice, we use a strong external model only as a \emph{verifier} that checks and normalizes answers, and we apply it identically to every method we compare. This requires addressing two challenges at once: \textbf{(1)} diagnosing model collapse at a granularity that is actionable for data curation, and \textbf{(2)} designing synthetic data construction strategies that support sustained model improvement.

To address the first question, we empirically examine how synthetic data changes the \emph{skill profile} of evolving models. Using GSM8K~\citep{cobbe2021gsm8k} as an illustrative case, we apply the Deterministic Inputs Noisy And gate model (DINA)~\citep{delatorre2009dina} from the family of cognitive assessment models in education science to estimate \emph{skill mastery} for both the base model and the fine-tuned model trained with accumulated synthetic data. Figure~\ref{fig:radar_skill} visualizes a subset of resulting skill profiles (full results on other datasets and models, fine-tuning settings, and data synthesis procedures are detailed in the Appendix~\ref{additional_experiments}.). We observe from Figure~\ref{fig:radar_skill}, on the one hand, synthetic data improves several skills that were previously strong enough, including \textit{Data} and \textit{Money \& Finance}. On the other hand, some weak skills not fully mastered by the base model become even weaker in the fine-tuned model, such as \textit{Algebra} and \textit{Time \& Scheduling}. This result suggests that model collapse in instruction tuning is not simply a uniform degradation of overall performance, but a \emph{polarization of competence}: synthetic data may further reinforce already strong capabilities while degrading weakly supported skills.

Motivated by the above finding, we address the second question by proposing \textbf{K}nowledge-boundary \textbf{I}nstruction \textbf{T}uning via \textbf{E}xploration (KITE), a two-stage framework for synthetic data construction in iterative model evolution. Overall, KITE is designed to provide useful learning signal so that newly added synthetic data improves rather than destabilizes the next-generation model. In the first stage, KITE constructs a large candidate bank by extracting weakness profiles from model failures guided by the DINA model, and then generating new data with rank-based noise injection, which encourages exploration beyond dominant high-probability regions. In the second stage, it curates this bank using Kernel Boundary Uncertainty (KBU), a likelihood-weighted semantic uncertainty score that identifies data near the model's semantic knowledge boundary. Experiments for instruction tuning on several reasoning datasets show that KITE enables stable and effective model evolution.

In summary, we make the following key contributions: \textbf{1)} we study model collapse in iterative synthetic data instruction tuning, and show that the failure is not uniform degradation but a \emph{polarization of
competence}; \textbf{2)} we propose \textbf{KITE}, whose two stages directly reverse the two arms of polarization via failure-guided generation and boundary-aware KBU curation. and empirically show across several datasets and multiple open-source LLMs that it yields more stable multi-generation improvement than strong synthetic data baselines.

\section{Related Work}
\label{related_work}

\subsection{Model Collapse}

Model collapse studies the failure mode that arises when later-generation models are trained on data increasingly contaminated by outputs from earlier generations. Early work by \citet{shumailov2024curserecursiontraininggenerated} shows that recursive training on model-generated data can progressively distort the learned distribution and erase low-probability modes, while \citet{dohmatob2024a} analyze the same phenomenon through the lens of scaling laws and argue that the loss of tail support leads to a cutoff in continued improvement, and further show that collapse can persist as long as the synthetic fraction does not vanish~\citep{dohmatob2025strong}. \citet{gerstgrasser2024is} and \citet{kazdan2025collapse} show that collapse is not inevitable when synthetic data are accumulated together with the original real data, in which case the test error can remain bounded. \citet{zhu2025how} propose token-level editing to construct semi-synthetic text. \cite{feng2025beyond} argue that scaling with synthesized data requires verification, showing that verifier-based selection can prevent collapse; \citet{drayson-etal-2025-machine} show that machine-generated text detection can be used to reduce collapse; and \citet{amin2025escaping} show that even weak but properly curated data can support continued model improvement. However, most existing work treats the problem primarily from the perspective of degeneration control, where the goal is to keep degradation bounded. In contrast, we study iterative model evolution for instruction tuning, where the key challenge is to diagnose collapse at a granularity that can guide data curation and support continued improvement from one generation to the next.

\subsection{LLMs for Synthetic Data}

Large language models have become a powerful tool for synthetic data generation~\citep{liu2024best}, substantially extending earlier approaches based on conventional language models~\citep{schick-schutze-2021-generating}. Recent work has used LLMs to construct synthetic data for a wide range of purposes, including multilingual question answering~\citep{riabi-etal-2021-synthetic}, dialogue systems~\citep{zhao2024wildchat}, instruction tuning~\citep{xu2025magpie, zhang2025oasisimageneedmultimodal, zhong2025synthet2cgeneratingsyntheticdata}, factuality improvement~\citep{wei2024simplesyntheticdatareduces}, scientific reasoning~\citep{huang2025chemorch}, and dataset diversification~\citep{dai2025auggpt, Riaz_2025}. More recent frameworks such as DataGen~\citep{huang2025datagen} and Janus~\citep{lee2024aligning} further highlight the potential of LLMs to produce high-quality synthetic corpora for model training, evaluation, and alignment. However, these works study synthetic data primarily as a means to improve downstream performance, often in settings where data are generated by stronger external models and used in a single-round training pipeline. This is different from the model collapse phenomenon, where later-generation models are trained on data generated from their own distribution. We focus on iterative model evolution under self-generated data.

\begin{figure*}[t]
    \centering
    \includegraphics[width=0.95\textwidth]{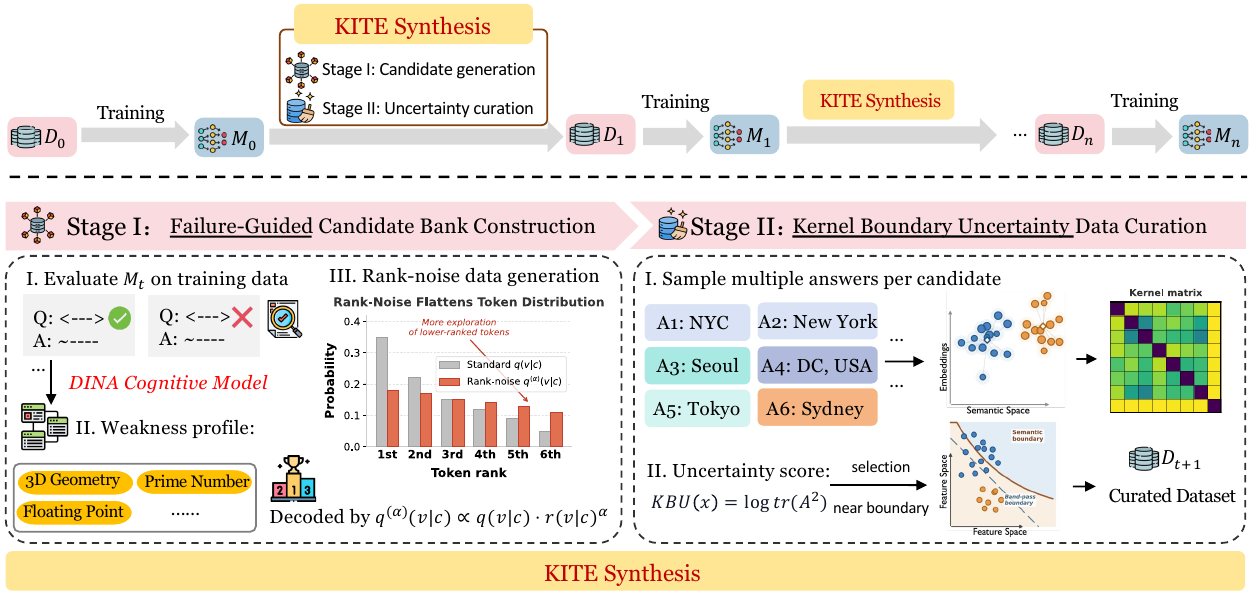}
    \caption{Top: iterative model evolution with KITE synthesis. Bottom: the two stages of KITE, failure-guided candidate bank construction and Kernel Boundary Uncertainty (KBU) data curation.}
    \label{fig:kite}
\end{figure*}

\section{Methodology}
\label{methodology}

\subsection{Notation and Formulation}
\label{sec:notation}

Let $M_t$ denote the instruction-following model at evolution step $t$ with output distribution $p_{M_t}(y\mid x)$, where $x$ is an instruction and $y$ is an answer. We start from an initial verified training set $D_0^\star=\{(x_i,y_i^\star)\}_{i=1}^{n_0}$ (e.g., human data) and mimic real-world model evolution by \emph{accumulating} training data over time: at step $t$, our pipeline produces a new synthetic curated set $D_t=\{(x_i)\}_{i=1}^{B}$ under a labeling budget $B$, which is then labeled to $D_t^\star=\{(x_i,y_i^\star)\}_{i=1}^{B}$, and the next model is trained on the union of all verified data so far under the supervised fine-tuning (SFT) objective. We emphasize that each $M_t$ is obtained by fine-tuning an identical base model on $D_t^\star$, rather than from $M_{t-1}$.

Concretely, we first generate a large synthetic candidate bank $\mathcal{B}_t=\{x_i\}_{i=1}^{N}$ with $N\gg B$, select $D_t\subseteq \mathcal{B}_t$ with $|D_t|=B$, and apply a strong verified labeler to obtain $D_t^\star$ for training.

We define \emph{model collapse} in our setting as a \emph{degradation of held-out performance}. Our objective is to design the data-construction operator that generates and curates training data so that model evolution is \emph{self-improving}. Iterative model evolution and the KITE framework are shown in Figure~\ref{fig:kite}. The rest of this section specifies our two-stage pipeline for constructing $\mathcal{B}_t$ and selecting $D_t$ so that each newly added slice $D_t^\star$ is both (i) targeted at current weaknesses and (ii) reliably improves the next model in the accumulated training regime.


\subsection{Stage 1: Failure-Guided Candidate Bank Construction}
\label{sec:candidate_bank}

At evolution step $t$, our first stage constructs a large synthetic \emph{candidate bank}
$\mathcal{B}_t$ of size $N\gg B$ whose role is to (i) \textit{expand coverage} beyond the model's internal
high-probability modes and (ii) concentrate generation on \emph{current weaknesses} so that
subsequent curation can spend verifier budget on examples with real learning value.

Concretely, we generate $\mathcal{B}_t$ by (1) extracting a weakness profile from training failures guided by the DINA model and (2) sampling new instructions with rank-based noise injection.



\paragraph{1) Failure-guided weakness profiling.}
A core driver of synthetic collapse is \emph{self-reinforcement}: unconstrained generation tends to
over-sample skills the model already performs well on, so the synthetic distribution drifts toward
easy regions and provides diminishing learning signal. To counter this drift, we explicitly tie
generation to \emph{observed deficiencies} of the current model.

Rather than using raw failed examples directly, we build a diagnosis pipeline based on the DINA
model from cognitive science (details in Appendix~\ref{diagnosis}). DINA is useful here because it
does not treat each failure as an isolated instance, instead, it infers which latent skills are likely
under-mastered from the model's response pattern over many examples. This gives a more structured
weakness estimate than simple example failure counting.

Concretely, each training instance is associated with a binary skill requirement vector from a
Q-matrix, indicating which fine-grained skills are needed to solve it. In our implementation, the
Q-matrix is constructed by LLM-assisted skill tagging followed by consistency
checking, while the full procedure is deferred to Appendix~\ref{diagnosis}. Given the model's
correct and incorrect responses on the current training set, DINA estimates posterior mastery for each
skill, and we convert low-mastery skills into short natural-language \emph{weakness descriptors}
(e.g., missing reasoning type or domain knowledge). We aggregate these descriptors into a weakness
profile $S_t=\{s_k\}_{k=1}^{K}$, and use each $s\in S_t$ to instantiate a question-generation prompt
template $\pi_Q(s)$ that asks for new instructions specifically targeting that weakness.

\paragraph{2) Rank-based noise injection.}
Even under weakness conditioning, standard decoding often produces easy or templated instructions.
A natural baseline is to increase sampling temperature. However, in our experiments, temperature tuning improves surface diversity but does \emph{not} reliably yield informative training signal. Related supporting experiment results are detailed in the Section~\ref{general}. We therefore use a rank-based perturbation that explicitly flattens the
\emph{ordering} of token probabilities.

Let $\ell_t(c)\in\mathbb{R}^{|V|}$ be the logits of $M_t$ at decoding context $c$ and
$q_t(v\mid c)=\mathrm{softmax}(\ell_t(c))_v$. Define the rank of token $v$ by its logit order,
$r_t(v\mid c)=1+\big|\{u:\ell_{t,u}(c)>\ell_{t,v}(c)\}\big|$. We then sample from the modified
distribution
\begin{equation}
\begin{aligned}
q^{(\alpha)}_t(v\mid c)
&= \operatorname{softmax}\!\Bigl(
    \ell_t(c)
    + \alpha \log r_t(\cdot\mid c)
   \Bigr)_v \\
&\propto q_t(v\mid c)\,
          r_t(v\mid c)^{\alpha}.
\end{aligned}
\label{eq:rank_noise}
\end{equation}
which smoothly upweights lower-ranked tokens and encourages exploration without hard truncation.
Empirically, we find that rank-noise decoding produces candidate instructions with more useful learning signal for subsequent training, which in turn leads to more stable self-improvement.

For each $s\in S_t$, we generate candidate instructions by decoding under
$q^{(\alpha_Q)}_t(\cdot\mid \pi_Q(s))$ and collect $\mathcal{B}_t=\{x_i\}_{i=1}^{N}$ by repeating across
descriptors until reaching total bank size $N$.

\subsection{Stage 2: Uncertainty-Based Data Curation}
\label{sec:uncertainty_curation}

Given the candidate bank $\mathcal{B}_t$, Stage~2 selects a budgeted subset
$D_t\subseteq \mathcal{B}_t$ with $|D_t|=B$ for labeling and training. The key idea is to
prioritize candidates that provide \emph{useful learning signal}: not examples that are too
easy for $M_t$, and not examples that fall far outside the model's knowledge boundary, where failure
is primarily due to missing capabilities inherited from pre-training and thus is less amenable to
post-training correction, but examples that lie near the model's \emph{semantic knowledge boundary}, where multiple plausible answers coexist. Motivated by the growing use of uncertainty to characterize the model's capacity, and in particular by Kernel Language Entropy (KLE) for hallucination detection~\citep{nikitin2024kernel}, we instantiate this intuition through a semantic uncertainty score over multiple sampled answers.

\paragraph{Semantic uncertainty as curation signal.}
Token-level uncertainty is often too local for instruction tuning, since it can be inflated by
surface variation or paraphrasing without reflecting genuine uncertainty about the underlying answer.
Instead, for each candidate instruction $x\in\mathcal{B}_t$, we probe $M_t$ by sampling
$m$ answers $y^{(1)},\dots,y^{(m)}\sim M_t(\cdot\mid x)$ and measure uncertainty in
\emph{semantic space}. Let $h_i=\phi(y^{(i)})\in\mathbb{R}^d$ be the embedding of answer $y^{(i)}$,
and define a Radial Basis Function (RBF) kernel
\begin{equation}
\begin{aligned}
K_{ij}
&= \exp\!\left(
    -\frac{\lVert h_i-h_j\rVert^2}{2\sigma^2}
   \right), \\
\sigma^2
&= \operatorname{median}
   \left\{
     \lVert h_i-h_j\rVert^2 : i<j
   \right\}.
\end{aligned}
\end{equation}
This kernel captures whether sampled answers belong to the same semantic mode or to distinct modes.

\paragraph{Boundary-aware uncertainty via likelihood weighting.}
A purely semantic dispersion score is not sufficient by itself: under stochastic decoding, some sampled
answers may be semantically distinct simply because they are low-probability off-manifold generations.
Treating all samples equally would therefore overestimate uncertainty and bias selection toward prompts
that are noisy or far outside the model's actionable knowledge boundary. For post-training data
curation, we want disagreement among \emph{plausible} answers, since that is more indicative of a prompt lying near the model's semantic knowledge boundary and thus more likely to yield useful learning signal.

To instantiate this idea, we weight each sampled answer by its likelihood under $M_t$. For a
candidate instruction $x$, let $y^{(1)},\dots,y^{(m)}\sim M_t(\cdot\mid x)$ be $m$ sampled answers.
We compute the length-normalized log-likelihood of each sample:
\begin{equation}
\begin{aligned}
\ell_i
&= \frac{1}{\lvert y^{(i)}\rvert}
   \sum_k
   \log p_{M_t}\!\left(
      y_k^{(i)}
      \,\middle|\,
      x, y_{<k}^{(i)}
   \right).
\end{aligned}
\end{equation}
We then convert these scores into normalized weights:
\begin{equation}
\tilde{w}_i
= \frac{\exp(\gamma \ell_i)}
       {\displaystyle\sum_j \exp(\gamma \ell_j)}.
\end{equation}
This weighting suppresses low-probability generations that would otherwise inflate uncertainty, while
preserving disagreement among answers that the model itself assigns substantial probability mass.

We then combine likelihood weighting with the semantic kernel defined above. This kernelized
formulation avoids making hard clustering decisions over sampled answers, and instead captures
uncertainty through their continuous semantic similarity structure. Let
$W=\mathrm{diag}(\sqrt{\tilde w_1},\dots,\sqrt{\tilde w_m})$, and form the unit-trace matrix
\begin{equation}
A
= \frac{WKW}
       {\operatorname{tr}(WKW)}.
\end{equation}
We define the \emph{Kernel Boundary Uncertainty} (KBU)
score of instruction \(x\) as the Rényi-2 entropy of \(A\):
\begin{equation}
U_{\mathrm{KBU}}(x)
= -\log \operatorname{tr}\!\left(A^2\right).
\end{equation}
We choose Rényi-2 entropy because it provides a stable measure of the effective number of
semantic modes represented in the weighted kernel, while remaining computationally lightweight.

When the KBU score is small, the sampled answers concentrate on a single semantic mode. This
typically indicates that $x$ is either too easy for the model, or the model is confidently wrong, but in either case it offers little curation signal. Intermediate values indicate disagreement among
multiple plausible answers, suggesting that $x$ lies near the model's semantic knowledge boundary and
is likely to provide useful learning signal. Excessively large values, however, may indicate that the
prompt lies beyond the model's actionable knowledge boundary. We therefore select candidates whose uncertainty falls within a target percentile range $(u_{\min}, u_{\max})$, where $u_{\min},u_{\max}\in(0,1)$ denote lower and upper quantile thresholds.

\begin{table*}[t!]
    \centering
    \vspace{-1pt}
    \renewcommand{\arraystretch}{1}
    \setlength{\tabcolsep}{6pt}\vspace{-2pt}
    \resizebox{1.0\linewidth}{!}{%
    \begin{tabular}{l l c c c c c}
\toprule[1pt]
\multirow{2}{*}{\textbf{Model}} 
  & \multirow{2}{*}{\textbf{Method}} 
  & \multicolumn{4}{c}{\textbf{Dataset (\%)}} 
  & \multirow{2}{*}{\textbf{Average}} \\
\cmidrule(lr){3-6}
  &  & \textbf{GSM8K} & \textbf{MMLU-Pro} & \textbf{MATH} & \textbf{GPQA} & \\
\midrule

\rowcolor{gray!8}
\cellcolor{white}{\multirow{7}{*}[-0.1ex]{\centering\textbf{Qwen-3-4B-Instruct}}}
  & Initial Model                    & 93.63 $\pm$ 0.18 & 69.55 $\pm$ 0.25 & 80.80 $\pm$ 0.32 & 40.40 $\pm$ 0.65 & 71.10 \\
  & Human data                       & 94.28 $\pm$ 0.22 & 70.40 $\pm$ 0.31 & 81.60 $\pm$ 0.40 & 41.41 $\pm$ 0.71 & 71.92 \\
\rowcolor{gray!8}
\cellcolor{white}{}
  & Self-Instruct                    & 93.79 $\pm$ 0.28 & 70.12 $\pm$ 0.35 & 81.00 $\pm$ 0.45 & 40.90 $\pm$ 0.81 & 71.45 \\
  & Few-shot syn.                    & 93.58 $\pm$ 0.30 & 70.01 $\pm$ 0.38 & 80.80 $\pm$ 0.42 & 40.40 $\pm$ 0.76 & 71.20 \\
\rowcolor{gray!8}
\cellcolor{white}{}
  & CDS                              & 94.03 $\pm$ 0.25 & 70.50 $\pm$ 0.29 & 81.20 $\pm$ 0.38 & 41.41 $\pm$ 0.68 & 71.79 \\
  & ToEdit                           & 94.50 $\pm$ 0.24 & 70.40 $\pm$ 0.33 & 81.60 $\pm$ 0.41 & 41.41 $\pm$ 0.74 & 71.98 \\
\rowcolor{gray!8}
\cellcolor{white}{}
  & Ours                             & \textbf{95.04 $\pm$ 0.19} & \textbf{71.87 $\pm$ 0.26} & \textbf{82.60 $\pm$ 0.35} & \textbf{41.92 $\pm$ 0.66} & \textbf{72.86} \\
\midrule

\rowcolor{blue!5}
\cellcolor{white}{\multirow{7}{*}[-0.1ex]{\centering\textbf{Qwen-3-1.7B}}}
  & Initial Model                    & 85.97 $\pm$ 0.22 & 51.10 $\pm$ 0.31 & 72.80 $\pm$ 0.35 & 28.28 $\pm$ 0.75 & 59.54 \\
  & Human data                       & 87.50 $\pm$ 0.28 & 52.49 $\pm$ 0.32 & 74.00 $\pm$ 0.42 & 28.79 $\pm$ 0.73 & 60.70 \\
\rowcolor{blue!5}
\cellcolor{white}{}
  & Self-Instruct                    & 87.01 $\pm$ 0.29 & 51.97 $\pm$ 0.26 & 73.40 $\pm$ 0.47 & 27.78 $\pm$ 0.82 & 60.04 \\
  & Few-shot syn.                    & 86.73 $\pm$ 0.31 & 52.40 $\pm$ 0.39 & 73.00 $\pm$ 0.44 & 27.78 $\pm$ 0.78 & 59.98 \\
\rowcolor{blue!5}
\cellcolor{white}{}
  & CDS                              & 87.04 $\pm$ 0.26 & 52.88 $\pm$ 0.30 & 74.40 $\pm$ 0.40 & \textbf{29.29 $\pm$ 0.69} & 60.90 \\
  & ToEdit                           & 87.50 $\pm$ 0.25 & 52.49 $\pm$ 0.34 & 74.00 $\pm$ 0.43 & 28.79 $\pm$ 0.75 & 60.70 \\
\rowcolor{blue!5}
\cellcolor{white}{}
  & Ours                             & \textbf{88.48 $\pm$ 0.27} & \textbf{53.30 $\pm$ 0.29} & \textbf{75.00 $\pm$ 0.38} & \textbf{29.29 $\pm$ 0.71} & \textbf{61.52} \\
\midrule

\rowcolor{gray!8}
\cellcolor{white}{\multirow{7}{*}[-0.1ex]{\centering\textbf{Llama-3.2-3B-Instruct}}}
  & Initial Model                    & 75.28 $\pm$ 0.30 & 38.11 $\pm$ 0.34 & 41.60 $\pm$ 0.45 & 30.30 $\pm$ 0.82 & 46.32 \\
  & Human data                       & 76.97 $\pm$ 0.28 & 39.03 $\pm$ 0.33 & 42.60 $\pm$ 0.49 & 31.31 $\pm$ 0.74 & 47.48 \\
\rowcolor{gray!8}
\cellcolor{white}{}
  & Self-Instruct                    & 76.18 $\pm$ 0.35 & 38.20 $\pm$ 0.41 & 41.60 $\pm$ 0.56 & 30.80 $\pm$ 0.83 & 46.70 \\
  & Few-shot syn.                    & 75.90 $\pm$ 0.37 & 38.65 $\pm$ 0.43 & 41.80 $\pm$ 0.52 & 30.80 $\pm$ 0.80 & 46.79 \\
\rowcolor{gray!8}
\cellcolor{white}{}
  & CDS                              & 76.97 $\pm$ 0.30 & 39.15 $\pm$ 0.35 & 42.60 $\pm$ 0.46 & 31.81 $\pm$ 0.71 & 47.63 \\
  & ToEdit                           & 77.03 $\pm$ 0.29 & 39.03 $\pm$ 0.34 & 42.60 $\pm$ 0.47 & 32.32 $\pm$ 0.73 & 47.75 \\
\rowcolor{gray!8}
\cellcolor{white}{}
  & Ours                             & \textbf{78.30 $\pm$ 0.32} & \textbf{39.62 $\pm$ 0.38} & \textbf{43.00 $\pm$ 0.43} & \textbf{32.83 $\pm$ 0.68} & \textbf{48.44} \\
\midrule

\rowcolor{blue!5}
\cellcolor{white}{\multirow{7}{*}[-0.1ex]{\centering\textbf{Llama-3-8B-Instruct}}}
  & Initial Model                    & 78.32 $\pm$ 0.25 & 38.75 $\pm$ 0.29 & 24.40 $\pm$ 0.42 & 32.32 $\pm$ 0.75 & 43.45 \\
  & Human data                       & 79.98 $\pm$ 0.27 & 39.38 $\pm$ 0.32 & 26.00 $\pm$ 0.48 & 32.83 $\pm$ 0.72 & 44.55 \\
\rowcolor{blue!5}
\cellcolor{white}{}
  & Self-Instruct                    & 79.15 $\pm$ 0.34 & 39.11 $\pm$ 0.40 & 25.00 $\pm$ 0.55 & 32.32 $\pm$ 0.80 & 43.90 \\
  & Few-shot syn.                    & 77.97 $\pm$ 0.36 & 38.75 $\pm$ 0.42 & 25.60 $\pm$ 0.51 & 32.32 $\pm$ 0.78 & 43.66 \\
\rowcolor{blue!5}
\cellcolor{white}{}
  & CDS                              & 79.91 $\pm$ 0.29 & 39.35 $\pm$ 0.34 & 26.20 $\pm$ 0.45 & 32.83 $\pm$ 0.69 & 44.57 \\
  & ToEdit                           & 80.05 $\pm$ 0.28 & 39.38 $\pm$ 0.33 & 26.00 $\pm$ 0.46 & 32.83 $\pm$ 0.71 & 44.57 \\
\rowcolor{blue!5}
\cellcolor{white}{}
  & Ours                             & \textbf{81.12 $\pm$ 0.23} & \textbf{39.85 $\pm$ 0.28} & \textbf{26.60 $\pm$ 0.42} & \textbf{33.33 $\pm$ 0.64} & \textbf{45.23} \\
\midrule

\rowcolor{gray!8}
\cellcolor{white}{\multirow{7}{*}[-0.1ex]{\centering\textbf{Gemma-3-4B-Instruct}}}
  & Initial Model                    & 84.69 $\pm$ 0.20 & 48.30 $\pm$ 0.28 & 66.60 $\pm$ 0.35 & 29.29 $\pm$ 0.77 & 57.22 \\
  & Human data                       & 85.98 $\pm$ 0.21 & 49.50 $\pm$ 0.30 & 68.00 $\pm$ 0.41 & 30.30 $\pm$ 0.70 & 58.45 \\
\rowcolor{gray!8}
\cellcolor{white}{}
  & Self-Instruct                    & 85.20 $\pm$ 0.27 & 48.84 $\pm$ 0.34 & 67.40 $\pm$ 0.46 & 28.79 $\pm$ 0.80 & 57.56 \\
  & Few-shot syn.                    & 85.57 $\pm$ 0.29 & 48.84 $\pm$ 0.37 & 67.40 $\pm$ 0.43 & 28.79 $\pm$ 0.75 & 57.65 \\
\rowcolor{gray!8}
\cellcolor{white}{}
  & CDS                              & 85.25 $\pm$ 0.24 & 49.50 $\pm$ 0.28 & 68.00 $\pm$ 0.39 & \textbf{30.81 $\pm$ 0.67} & 58.39 \\
  & ToEdit                           & 85.98 $\pm$ 0.23 & 49.50 $\pm$ 0.32 & 68.00 $\pm$ 0.42 & 30.30 $\pm$ 0.73 & 58.44 \\
\rowcolor{gray!8}
\cellcolor{white}{}
  & Ours                             & \textbf{86.52 $\pm$ 0.25} & \textbf{50.01 $\pm$ 0.29} & \textbf{69.40 $\pm$ 0.38} & \textbf{30.81 $\pm$ 0.69} & \textbf{59.19} \\
\bottomrule[1pt]
\end{tabular}
    }
    \caption{\textbf{Main results across models and synthetic data construction methods.}
    We report evaluation accuracy (\%) on four reasoning benchmarks after five iterations of accumulated self-improvement, averaged over 3 random seeds.}
    \label{tab:main_results}
    \vspace{-10pt}
\end{table*}

\section{Experiments}
\label{experiments}

\subsection{Experimental Setting}

\paragraph{Dataset.}
To evaluate our method, we use four widely adopted reasoning datasets covering complementary forms of difficulty. \textbf{GSM8K}~\citep{cobbe2021gsm8k} tests grade-school arithmetic and multi-step numerical reasoning; \textbf{MMLU-Pro}~\citep{wang2024mmlupro} extends MMLU~\citep{hendrycks2021measuringmassivemultitasklanguage} across diverse academic subjects; \textbf{MATH}~\citep{hendrycks2021measuringmathematicalproblemsolving} contains competition-level problems requiring symbolic reasoning and longer derivations; and \textbf{GPQA}~\citep{rein2024gpqa} consists of graduate-level science questions difficult even for strong LLMs. Full evaluation protocols are in Appendix~\ref{additional_experiments}.

\paragraph{Baselines.}
We compare our method with baselines that adopt different synthetic data construction strategies for iterative self-improvement. For all synthetic-data baselines, we follow the same accumulated training protocol: at each iteration, new synthetic data are generated, labeled, appended to the training set, and a new model is trained from the same base model on the accumulated labeled data. We report results after five iterations. \textbf{Human data only} serves as a non-synthetic control, where training uses only the initial verified human dataset. \textbf{Self-Instruct}~\citep{wang2023selfinstructaligninglanguagemodels} generates new instruction--response pairs by prompting the model to bootstrap synthetic tasks from a small seed set. \textbf{Few-shot synthesis} generates synthetic instruction--response pairs by few-shot prompting from seed examples. \textbf{Cognitive Diagnosis-based Synthesis (CDS)}~\citep{zhao-etal-2025-cds} uses cognitive-diagnosis-derived weakness profiles to guide targeted data generation but omits the uncertainty-based curation stage. \textbf{Token-Level Editing (ToEdit)}~\citep{zhu2025how} produces semi-synthetic data by diversifying model-generated data through token-level perturbation and editing.

\paragraph{Others.}
Due to the space limitations, \emph{Training} and \emph{Evaluation} setting are detailed in the Appendix~\ref{additional_experiments}. \emph{Cost Analysis} and \emph{Theoretical Analysis} are detailed in the Appendix~\ref{cost} and~\ref{theory}

\subsection{Main Results}

Table~\ref{tab:main_results} summarizes the performance of different synthetic data construction strategies across datasets and backbone LLMs after five iterations of model evolution. Overall, KITE achieves the best average performance within every model group, demonstrating its effectiveness for synthetic data driven instruction tuning. For example, on Qwen-3-4B-Instruct, KITE reaches 95.04\% on GSM8K, 71.87\% on MMLU-Pro, and 82.60\% on MATH, consistently outperforming the corresponding baseline settings. We also note that the gains on MMLU-Pro and GPQA are generally smaller than those on GSM8K and MATH, which is expected since these benchmarks are more challenging and often require broader knowledge and reasoning capacity that are harder to improve through instruction tuning alone.

In contrast, other synthetic data baselines provide only limited gains. Because we retain all verified real data across iterations (the accumulate setting of \citet{gerstgrasser2024is}), catastrophic accuracy collapse is bounded by construction; the relevant failure mode here is therefore the inability of self-generated data to push the model past the human-data-only ceiling. This is exactly what the baselines exhibit. Self-Instruct and Few-shot avoid severe degradation relative to the Initial Model, but they remain \emph{below} the Human data setting. CDS performs better than generic synthesis but below KITE, indicating that weakness-guided generation alone is insufficient without uncertainty curation. ToEdit can improve over simpler baselines, but remains consistently below KITE, showing that increasing diversity at the generation stage does not by itself guarantee the most useful learning signal.

\subsection{Ablation Studies}

We modify KITE’s components to conduct an ablation study and validate each design choice. Table~\ref{tab:ablation} reports results on Llama-3-8B-Instruct. The full method achieves the best average performance at 45.2\%, compared with 43.2\% without weakness profiling and 43.9\% when KBU is replaced with token entropy.

To assess candidate-bank construction, we remove weakness profiling and rank-noise injection. The performance drop in \textbf{w/o weakness.} shows that targeting model deficiencies is important for generating useful synthetic data, while the drop in \textbf{w/o rank-noise.} suggests that sampling only from dominant model modes is insufficient to preserve coverage. Replacing KBU with token entropy also consistently underperforms the full method, supporting the use of \emph{semantic} rather than token-level uncertainty. Removing \textbf{likelihood} weighting further degrades performance, confirming that uncertainty should emphasize disagreement among \emph{plausible} answers rather than low-probability, off-manifold generations. We also vary band-pass curation range. Both \textbf{$u_{\min}=0.4,;u_{\max}=1.0$} and \textbf{$u_{\min}=0.0,;u_{\max}=0.6$} underperform the full method, indicating that selecting only high- or low-uncertainty examples is suboptimal. This supports our hypothesis that the most useful learning signal lies near an intermediate semantic knowledge boundary.

\begin{table}[t!]
\centering
\small
\setlength{\tabcolsep}{2pt}
\label{tab:ablation}
\rowcolors{2}{blue!3}{white}
\resizebox{\columnwidth}{!}{%
\begin{tabular}{lccccc}
\toprule[1pt]
\textbf{Ablation} & \textbf{GSM8K} & \textbf{MMLU-Pro} & \textbf{MATH} & \textbf{GPQA} & \textbf{Avg.} \\
\midrule
w/o weakness.            & 76.2 & 39.1 & 25.0 & 32.3 & 43.2 \\
w/o rank-noise.          & 80.4 & 39.9 & 26.2 & 33.3 & 44.9 \\
token entropy.        & 79.9 & 39.3 & 25.0 & 31.3 & 43.9 \\
w/o likelihood.          & 80.0 & 39.3 & 26.6 & 32.3 & 44.6 \\
(0.4,\;1.0)    & 79.7 & 39.5 & 26.2 & 33.3 & 44.7 \\
(0.0,\;0.6)   & 79.5 & 39.7 & 26.2 & 32.8 & 44.6 \\
\midrule
\rowcolor{green!3}
\textbf{Full method} & \textbf{81.1} & \textbf{39.9} & \textbf{26.6} & \textbf{33.3} & \textbf{45.2} \\
\bottomrule[1pt]
\end{tabular}%
}
\caption{\textbf{Ablation study on Llama-3-8B-Instruct.} We evaluate the contribution of each component of our method.}
\label{tab:ablation}
\vspace{-10pt}
\end{table}

\begin{figure*}[t]
    \centering
    \includegraphics[width=0.9\linewidth]{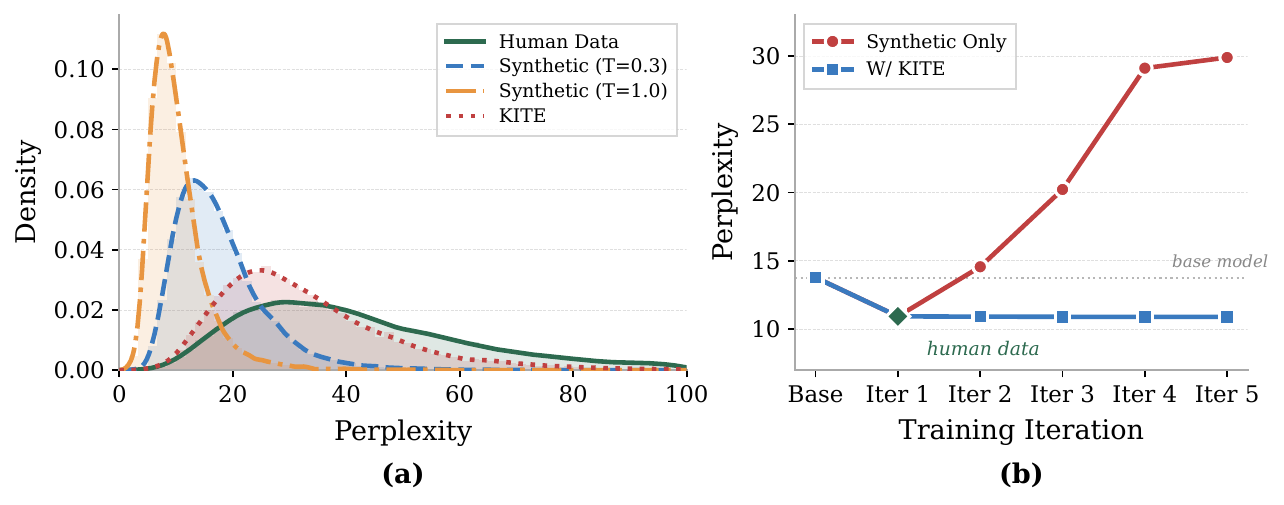}
    \caption{\textbf{Perplexity-based analysis of synthetic data under recursive training.} (a) Perplexity distributions on wikitext-2 with Llama-3-8B-Instruct for human-written data, synthetic data generated with temperature=0.3, synthetic data generated with temperature=1.0, and synthetic data generated with KITE. (b) Perplexity on the wikitext-2 test set across recursive model generations when Llama-3-8B-Instruct trained on synthetic data.}
    \label{fig:ppl}
\end{figure*}

\subsection{Long-Horizon Evolution Dynamics}

\begin{figure}
    \centering
    \includegraphics[width=\linewidth]{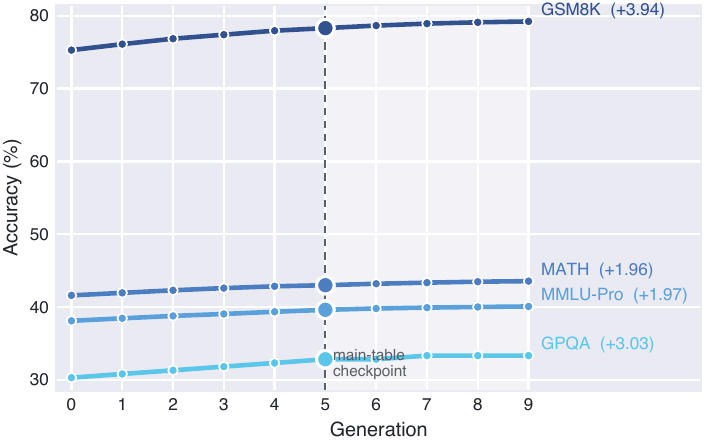}
    \caption{Long-horizon evolution dynamics of KITE on Llama-3.2-3B-Instruct.}
    \label{fig:long_horizon}
\end{figure}

To test whether KITE keeps improving beyond the five iterations of Table~\ref{tab:main_results} or
eventually stagnates or reverses, we extend Llama-3.2-3B-Instruct to nine generations under the
identical pipeline (Figure~\ref{fig:long_horizon}; Generation~0 is the initial model and
Generation~5 matches the ``Ours'' row). Performance increases \emph{monotonically} on every
benchmark with no degradation-and-reversal signature of collapse, and the gains \emph{saturate} gradually. This
is the expected behavior of a well-behaved self-improvement loop, continued but diminishing returns
rather than runaway gains or collapse.

\subsection{Generalization to Non-Reasoning Tasks}
\label{general}

To evaluate KITE beyond reasoning tasks, we follow \citet{shumailov2024curserecursiontraininggenerated} and convert wikitext-2~\citep{merity2016pointersentinelmixturemodels} into an instruction-tuning dataset for recursive self-training. Because this task has no explicit skill taxonomy, we apply KITE without weakness profiling. As shown in Figure~\ref{fig:ppl}(a), standard decoding at both low and high temperatures produces synthetic data overly concentrated in the low-perplexity region, whereas KITE yields a broader distribution closer to human-written data. Figure~\ref{fig:ppl}(b) further examines the consequence of this phenomenon under iterative training. When synthetic data generated with high-temperature are used for repeated self-training, the perplexity on the wikitext-2 test set increases monotonically from Model~1 to Model~5, which is aligned with insights discussed by \citet{shumailov2024curserecursiontraininggenerated}, while KITE keeps perplexity stable across iterations.

\begin{figure}
    \centering
    \includegraphics[width=\linewidth]{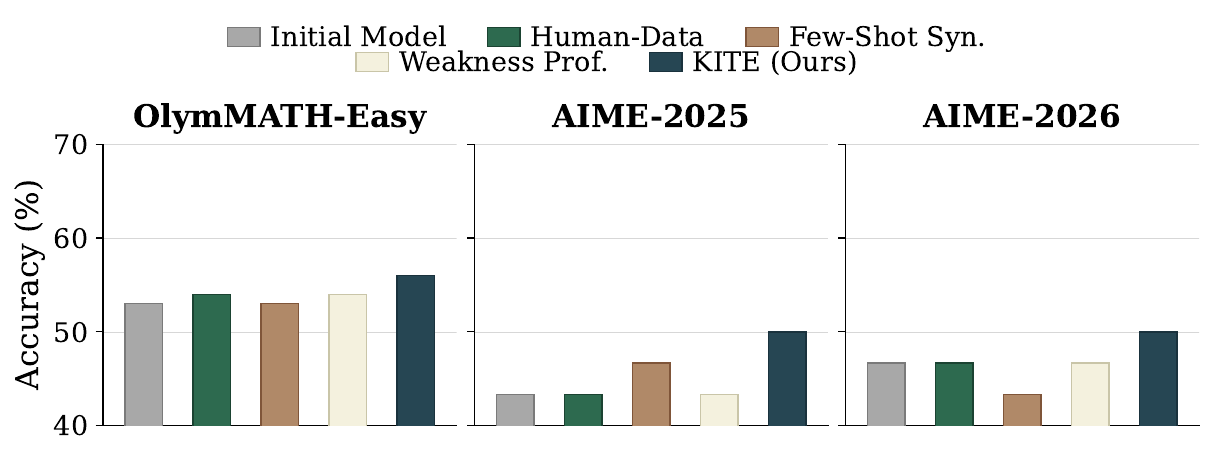}
    \caption{\textbf{OOD generalization on math reasoning benchmarks.} We evaluate Qwen-3-4B-Instruct-2507, trained with the above in-domain math subsets, on the unseen OlymMATH and AIME benchmarks.}
    \label{fig:ood}
\end{figure}

\subsection{OOD Generalization}

Figure~\ref{fig:ood} shows whether KITE trained on all math subsets of the four in-domain datasets described above can generalize to out-of-distribution (OOD) mathematical reasoning benchmarks. We evaluate Qwen-3-4B-Instruct on \textbf{OlymMATH}~\citep{sun2025challengingboundariesreasoningolympiadlevel} and \textbf{AIME}~\citep{aops_aime_problems}, harder competition-style problems never seen during training or synthetic data construction. KITE improves on both, suggesting that synthetic self-improvement enhances the model's underlying reasoning skills rather than merely fitting the benchmark distribution.

\section{Conclusion}

We study model collapse in synthetic data-driven instruction tuning under iterative model evolution, where the objective is to ensure that each successive model improves over its predecessor. We show that collapse can manifest as a polarization of competence, and propose KITE, a two-stage framework that combines failure-guided exploration with boundary-aware uncertainty curation. Experiments across multiple LLMs and benchmarks show that KITE yields more stable improvement than strong synthetic-data baselines, with ablations confirming the importance of both components.

\section{Limitations}

Our study has several limitations that suggest directions for future work. First, all methods
use a strong external model (gpt-5-mini) to verify and normalize answers; although it is applied
identically to KITE and every baseline and thus does not confound the relative comparison (Appendix~\ref{additional_experiments}), the
answer channel still carries an element of distillation, so our contribution is best read as a
study of self-generated \emph{instruction} distributions rather than pure closed-loop
self-improvement. Second, our DINA-based weakness profiling relies on an LLM-tagged,
non-expert-validated Q-matrix with fixed slip and guess priors, so the mastery scores are a
structured heuristic for steering generation rather than calibrated measurements, and we leave a
per-skill post-KITE re-diagnosis to future work. Finally, our evaluation focuses on
reasoning-style instruction tuning and non-reasoning perplexity analysis with $1.7$--$8$B backbones; scaling KITE to larger models, RL-style post-training, and a broader range of tasks is an important next step.

\section{Ethical Considerations}
This work studies data-construction methods for instruction tuning and does not introduce
new human-subject data collection. All datasets used are publicly released research benchmarks used in accordance
with their licenses and intended research use. The synthetic instructions do not target individuals or sensitive
attributes. Our use of an LLM as labeler and as evaluation judge can propagate biases or
errors present in that model; we mitigate this by preferring deterministic answer
extraction and exact-match whenever possible and restricting the judge to
formatting/equivalence decisions (Appendix~\ref{additional_experiments}), but residual
judge bias cannot be fully excluded.


\bibliography{custom}

\clearpage
\section{Appendix}

\subsection{Additional Related Works}

\paragraph{Self-improving LLMs.}
A growing body of work improves LLMs from their own generations. \citet{zelikman2022starbootstrappingreasoningreasoning} propose STaR, which bootstraps reasoning by iteratively generating rationales, retaining those that lead to correct answers, and fine-tuning on them. \citet{chen2024selfplayfinetuningconvertsweak} propose self-play fine-tuning (SPIN), where a model improves by discriminating its own responses against human-annotated data. \citet{gulcehre2023reinforced} introduce Reinforced Self-Training (ReST), which grows a dataset by sampling from the policy and filtering with a reward model, and \citet{yuan2024selfrewarding} propose Self-Rewarding Language Models, where the model acts as its own judge to supply training rewards. More recently, \citet{wu2026sgalm} formulate alignment as a self-generative adversarial game within a single LLM that doubles as a synthetic-data engine, while \citet{roe2026idempotent} show that iteratively fine-tuning a model on its own outputs is \emph{mostly idempotent}, with behavioral traits stabilizing rather than amplifying, an observation consistent with the diminishing-returns saturation we report over nine generations (Figure~\ref{fig:long_horizon}). These works establish that self-generated supervision can drive improvement, but they primarily introduce new \emph{training paradigms} (rationale bootstrapping, self-play, reward-filtered RL, adversarial games). In contrast, our work is not a new training objective but a synthetic-data generation and curation framework for iterative model evolution: we focus on how to construct the self-generated \emph{instruction} distribution so that later-generation models can keep improving without collapse, and KITE is orthogonal to and composable with these paradigms.

\paragraph{Synthetic data and data curation.}
Our second stage is closely related to work on weakness profiling and data selection. \citet{zhao-etal-2025-cds} propose CDS, which uses Cognitive Diagnosis Theory to characterize model weaknesses at the knowledge-component level and synthesize targeted data, closely aligned with our DINA-based diagnosis; we include CDS as a baseline in Table~\ref{tab:main_results}. \citet{sobhani-etal-2025-categorize} show that LLMs can categorize system inputs into fine-grained, natural-language performance categories, and \citet{zeng2025evaltree} build hierarchical capability trees (EvalTree) to profile model weaknesses more precisely; both provide structured alternatives for identifying where a model underperforms, complementary to our psychometric weakness estimate. On the curation side, \citet{xia2024less} (LESS) select influential data via low-rank gradient similarity to a target capability, \citet{li2024superfiltering} (Superfiltering) use a small proxy model's instruction-following-difficulty scores to filter data efficiently, and \citet{yang2025dynamic} study dynamic data selection jointly with data augmentation to accelerate training. Very recent work continues this thread: \citet{fan2026optimsyn} (Optimsyn) use gradient-based influence estimation to optimize synthetic-data generation rubrics via reinforcement learning, and \citet{amin2026erm} analyze the limitations of empirical risk minimization when training on mixed natural and synthetic data, showing that appropriate reweighting rather than plain ERM is needed to avoid collapse. These methods largely select from, reweight, or optimize generation over a pool using influence, difficulty, or quality heuristics. KITE differs in two ways: it \emph{generates} candidates targeted at diagnosed weaknesses rather than only filtering or reweighting existing data, and its Kernel Boundary Uncertainty (KBU) selects examples near the model's \emph{semantic knowledge boundary}, giving an explicit anti-collapse curation mechanism tailored to iterative synthetic-data self-improvement.

\subsection{Detailed Experiments Setting and Additional Results}
\label{additional_experiments}

\subsubsection{Models}
We evaluate our method on five open-source instruction-tuned LLMs spanning different model families and parameter scales: \textbf{Qwen3-4B-Instruct-2507}, \textbf{Qwen-3-1.7B}~\citep{qwen3technicalreport}, \textbf{Llama-3.2-3B-Instruct}, \textbf{Llama-3-8B-Instruct}~\citep{llama3modelcard}, and \textbf{Gemma-3-4B-It}~\citep{gemma_2025}. Unless otherwise specified, each model serves as the common initialization point for all compared methods, and every evolved model is trained from the same initial checkpoint under the corresponding accumulated data setting.

\subsubsection{Datasets}
We evaluate on \textbf{GSM8K}, \textbf{MMLU-Pro}, \textbf{MATH}, and \textbf{GPQA}. GSM8K is used in its standard form. For \textbf{MMLU-Pro}, we include only the science- and mathematics-related subject areas, and exclude domains such as law, business, and philosophy. For \textbf{MATH}, following prior practice, we construct a 500-example evaluation subset by uniform random sampling from the benchmark. For \textbf{GPQA}, we evaluate only on the \textbf{GPQA-Diamond} split. These choices allow us to focus evaluation on challenging reasoning tasks while keeping the benchmark suite computationally manageable and aligned with the goals of this work. Example instances for each dataset are included in Table~\ref{tab:data_examples}.

\subsubsection{Baselines}
Our baseline set is chosen to isolate the axis we study, \emph{how synthetic instructions are
constructed and selected}, and therefore holds the training objective fixed (accumulated SFT) across
all methods. We deliberately do \emph{not} treat self-improvement \emph{training paradigms} such as
STaR~\citep{zelikman2022starbootstrappingreasoningreasoning} or SPIN~\citep{chen2024selfplayfinetuningconvertsweak}
as competitors: they modify the learning objective (rationale bootstrapping, self-play discrimination)
rather than the instruction distribution, and KITE's data-construction operator is orthogonal to and
composable with them. Comparing KITE against a different training paradigm would confound the two
axes; instead, we compare against the data-construction strategies (generic synthesis, weakness
targeting, and semi-synthetic editing) that are directly substitutable for our operator under an
identical training pipeline. ToEdit and CDS in particular are strong, recent points of
comparison for the diversification and targeting intuitions, respectively.

\subsubsection{Training setting}
We use LLaMA-Factory~\citep{zheng-etal-2024-llamafactory} as the training framework for all supervised fine-tuning experiments. Unless otherwise specified, we fine-tune each model with LoRA under the SFT stage, using \texttt{lora\_rank}=8. For optimization, we use a learning rate of $1\times 10^{-5}$, a warmup ratio of 0.1, and train for 1 epoch. All experiments are conducted in \texttt{bf16} precision. Other implementation details follow the default LLaMA-Factory settings unless explicitly noted.

The initial verified set $D_0^\star$ is the human-authored training split of each dataset,
held strictly disjoint from the held-out evaluation instances reported in
Table~\ref{tab:main_results}; for benchmarks without a large native training split we seed
$D_0^\star$ from the released development/auxiliary items and never draw seeds from the
evaluation split. The ``Human data'' control reuses only $D_0^\star$ across all iterations,
whereas every synthetic method adds $B$ newly verified instructions per iteration on top of
$D_0^\star$, so all methods see the same real data and differ only in the synthetic
additions.

For KITE-specific data construction, we use rank-noise decoding with question-generation strength $\alpha_Q=0.75$, candidate-bank size $N=2,000$, and labeling budget $B=500$ per iteration. In Stage~2, we sample $m=10$ answers per candidate, use likelihood-weighting temperature $\gamma=0.8$, and select candidates by KBU quantile thresholds $(u_{\min},u_{\max})=(0.2,0.8)$. For DINA-based weakness profiling, we select weak skills using the mastery threshold $\tau=0.30$. The semantic embedding function $\phi(\cdot)$ used to construct the kernel is text-embedding-3-small~\citep{openai2026gpt5}. Unless otherwise noted, these hyperparameters are shared across all datasets and model backbones. We use an external labeler to verify correct candidate answers before training. Concretely, we use gpt-5-mini~\citep{openai2026gpt5} as the verifier. The detailed prompt template is provided in the Appendix~\ref{prompt}.

\subsubsection{Evaluation}
We evaluate model performance using an LLM-as-judge protocol~\citep{zheng2023judgingllmasajudgemtbenchchatbot}, where the judged model output is compared against the ground-truth answer to determine correctness. Concretely, we use gpt-5-mini~\citep{openai2026gpt5} as the judge model. We first apply deterministic answer extraction and exact-match comparison whenever possible, and use the judge only to handle semantically equivalent variants that differ in formatting or phrasing. The detailed evaluation prompt template is provided in Appendix~\ref{prompt}.

Our setting also let us rule out the most natural confound, that KITE's gains merely reflect
distillation from the external strong verifier. Self-Instruct, Few-shot synthesis, and ToEdit all consume the
\emph{identical} gpt-5-mini verification channel that KITE uses, yet several of them sit
\emph{below} the human-data ceiling (e.g., Self-Instruct and Few-shot on Qwen-3-1.7B). If verified
labels were the source of improvement, every method drawing on the same verifier would inherit it
equally; instead, only the method that changes \emph{which instructions are synthesized and selected}
moves above the ceiling. The verifier is thus better read as a shared, held-constant labeling
oracle than as the driver of the effect, which is the whole point of holding the answer channel
fixed. We return to the residual distillation caveat in the Limitations.

\subsubsection{Skill Profile}

\begin{figure*}[t]
    \centering
    \includegraphics[width=0.8\linewidth]{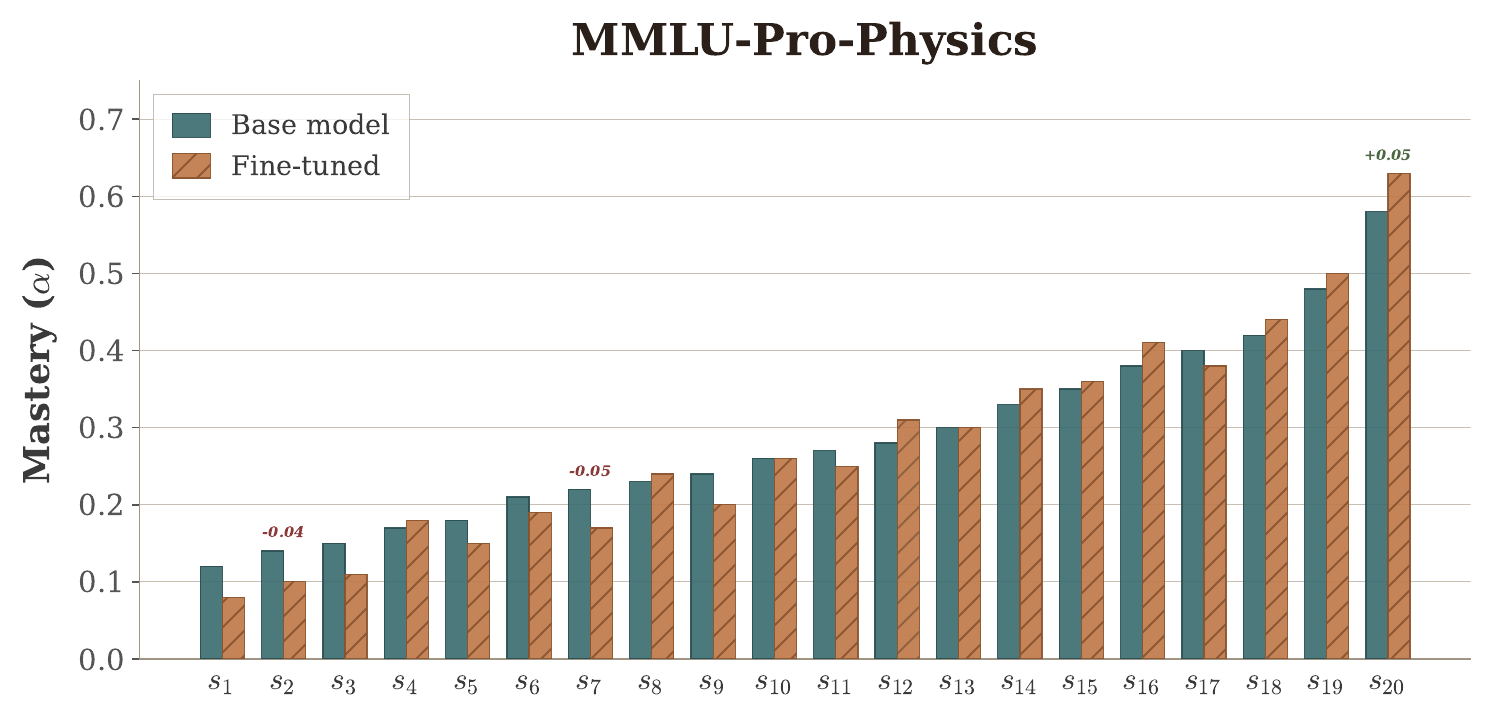}
    \caption{Skill mastery profile on MMLU-Pro-Physics estimated by DINA for the initial and fine-tuned models. Each $s_i$ denotes a dataset-specific skill defined in Table~\ref{tab:skill_index_compact}.}
    \label{fig:skill_mmlu}
\end{figure*}

\begin{figure*}[t]
    \centering
    \includegraphics[width=0.8\linewidth]{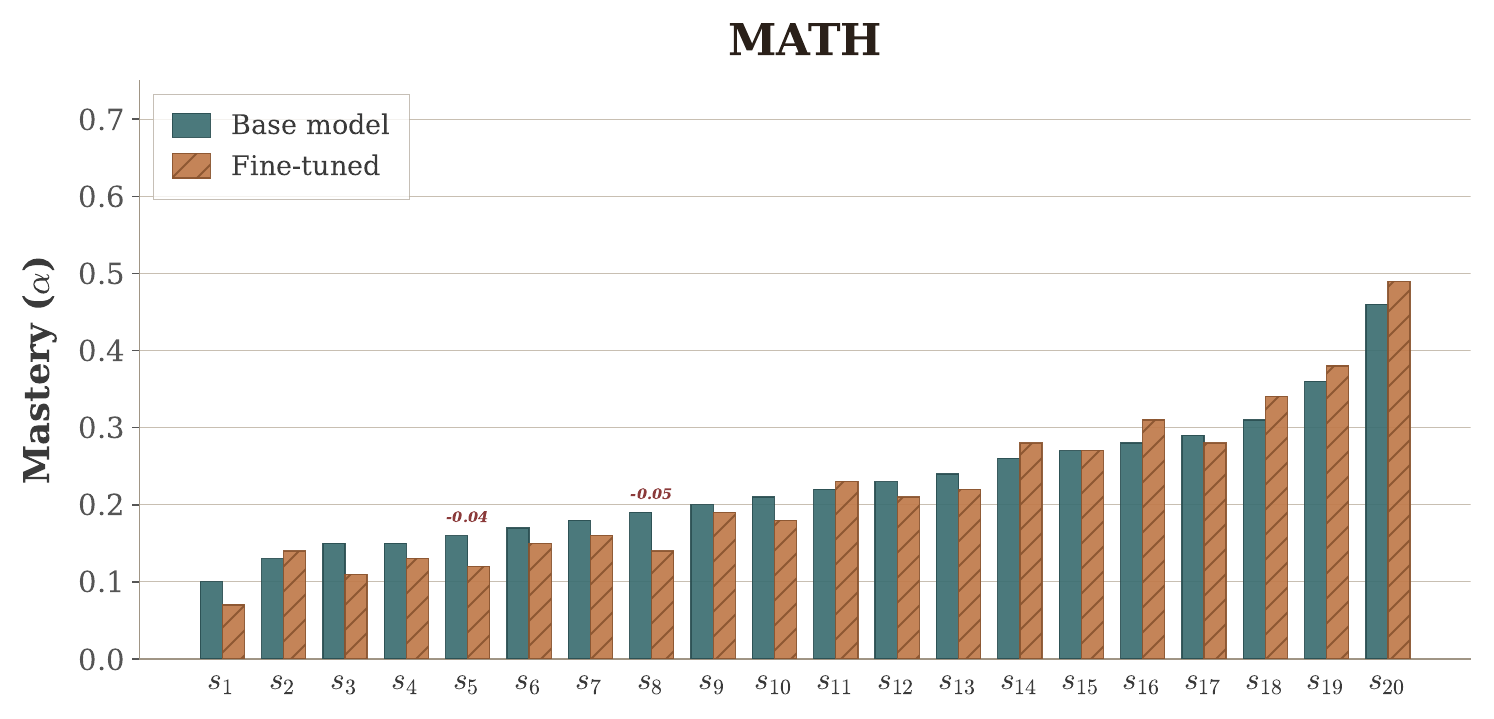}
    \caption{Skill mastery profile on MATH estimated by DINA for the initial and fine-tuned models. Each $s_i$ denotes a dataset-specific skill defined in Table~\ref{tab:skill_index_compact}.}
    \label{fig:skill_math}
\end{figure*}

\begin{figure*}[t]
    \centering
    \includegraphics[width=0.8\linewidth]{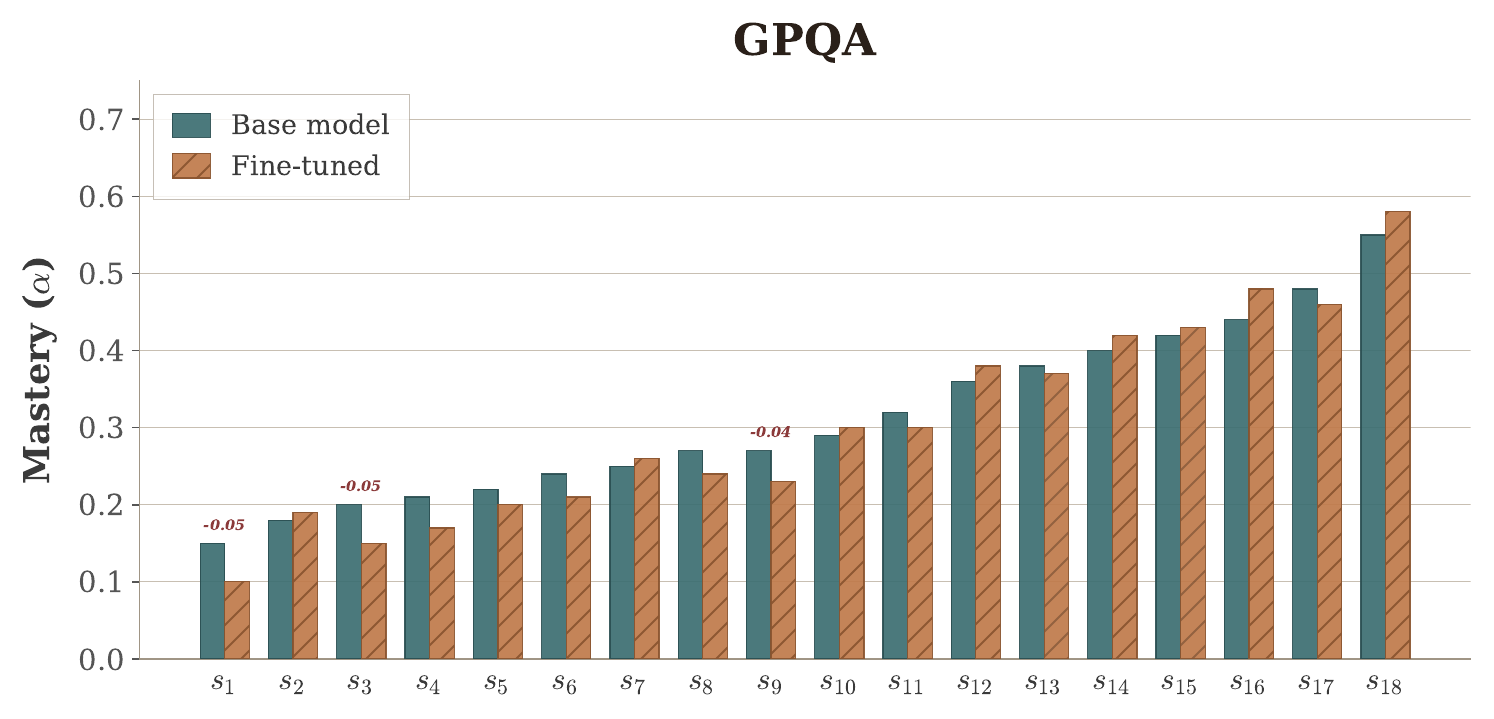}
    \caption{Skill mastery profile on GPQA estimated by DINA for the initial and fine-tuned models. Each $s_i$ denotes a dataset-specific skill defined in Table~\ref{tab:skill_index_compact}.}
    \label{fig:skill_gpqa}
\end{figure*}

We provide additional skill-profile results. For each dataset and model, we construct a diagnostic Q-matrix over fine-grained skills and fit the DINA model~\citep{delatorre2009dina} to the response patterns of the initial model and the corresponding fine-tuned model after accumulated synthetic training. The resulting posterior mastery scores allow us to compare how synthetic data changes skill-level competence beyond aggregate accuracy. Figures~\ref{fig:skill_mmlu}, \ref{fig:skill_math}, \ref{fig:skill_gpqa}, and Table~\ref{tab:skill_index_compact} show representative skill profiles for Llama-3-8B-Instruct on different datasets.

Across settings, we observe a consistent qualitative pattern. Synthetic training does not simply shift all skills upward or downward uniformly. Instead, it tends to further strengthen skills that are already relatively well supported, while weak or under-supported skills often improve little or even deteriorate. These results support our claim that collapse under synthetic data is better understood as a \emph{polarization of competence} rather than uniform degradation.

\subsection{Detailed Diagnosis Pipeline}
\label{diagnosis}

We diagnose the weakness profile of a target model $M_t$ using the DINA model from the cognitive diagnosis literature. DINA assumes that each evaluation example requires a subset of latent skills, and infers skill mastery from the model's pattern of correct and incorrect responses. We then convert low-mastery skills into the weakness profile used in Stage~1 of KITE.

Let $\mathcal{T}=\{(x_j,y_j^\star)\}_{j=1}^{J}$ be a diagnostic set, which is the training dataset in our implementation, and let $\mathcal{K}=\{1,\dots,K\}$ denote the skill inventory. Each example $j$ is associated with a binary Q-matrix row $q_j=(q_{j1},\dots,q_{jK})\in\{0,1\}^K$, where $q_{jk}=1$ indicates that item $j$ requires skill $k$. In practice, the Q-matrix is constructed from LLM-assisted skill tagging. For the target model $M_t$, we evaluate it on $\mathcal{T}$ and record a binary response vector $r^{(t)}\in\{0,1\}^{J}$, where
$
r_j^{(t)}=\mathbf{1}[\mathrm{Eval}(M_t(x_j),y_j^\star)=1].
$

Let $\alpha^{(t)}=(\alpha_1^{(t)},\dots,\alpha_K^{(t)})\in\{0,1\}^K$ denote the latent skill mastery vector of $M_t$, where $\alpha_k^{(t)}=1$ indicates mastery of skill $k$. Under DINA, an item can be solved ideally only if all required skills are mastered, with ideal response
$
\eta_j^{(t)}=\prod_{k=1}^{K}(\alpha_k^{(t)})^{q_{jk}}.
$ Each item $j$ is associated with a slip parameter $s_j$ and a guess parameter $g_j$, and DINA models the response probability as
$
\Pr(r_j^{(t)}=1 \mid \alpha^{(t)})=(1-s_j)^{\eta_j^{(t)}} g_j^{\,1-\eta_j^{(t)}}.
$
Unlike the standard psychometric setting, where slip/guess and the mastery distribution are
jointly estimated across a \emph{population} of respondents, we diagnose a \emph{single}
model. To keep the mastery estimate identifiable from one response vector, we do \emph{not}
free-estimate item parameters; instead we fix weak, shared slip/guess priors
($s_j=g_j=0.1$ unless noted) and a factorized skill prior, and compute the posterior over the
skill-mastery vector under the DINA likelihood, which factorizes over items. Because many
items share each skill through the Q-matrix, aggregating correct/incorrect responses across
all items requiring skill $k$ yields a stable posterior for that skill even from a single
respondent. We define the posterior mastery of skill $k$ as
$
m_k^{(t)}=\Pr(\alpha_k^{(t)}=1\mid r^{(t)}),
$
and the corresponding weakness score as
$
w_k^{(t)}=1-m_k^{(t)}.
$
This is a deliberate simplification of full DINA; as noted in the Limitations, we treat the
resulting scores as a structured heuristic for steering generation rather than as calibrated
latent-trait measurements, and KITE's downstream gains do not require the diagnosis to be
exact.

Finally, we form the weakness profile by selecting the most under-mastered skills, e.g., $\mathcal{W}_t=\{k:w_k^{(t)}\ge\tau\}$ or equivalently the top-$L$ skills ranked by $w_k^{(t)}$. These skills are then converted into natural-language weakness descriptors $S_t=\{s_1,\dots,s_L\}$, which are used to guide candidate generation in Stage~1. In this way, DINA provides a structured mechanism for translating fine-grained diagnosis results into targeted synthetic data construction.

\subsection{Cost Analysis}
\label{cost}

We briefly report the computational cost of KITE. For each model, a full 5-iteration run consists of three stages: synthetic candidate generation, KBU-based uncertainty curation, and supervised fine-tuning. Among them, the dominant cost comes from repeated fine-tuning, while the additional overhead of KITE mainly arises from Stage~II, where we sample multiple answers per candidate to estimate uncertainty. In our implementation, candidate generation and curation account for approximately 10\% and 12\% of the total wall-clock time, respectively, while supervised fine-tuning accounts for the remaining 78\%. Although KITE introduces extra synthesis and curation cost compared with simpler data synthesis baselines, the overhead remains affordable relative to training.

It is worth being explicit about the cost/benefit trade-off, since the per-iteration gains are
modest. All methods in our comparison pay the same dominant costs, the same labeling budget $B$ and
the same fine-tuning, so KITE's \emph{marginal} overhead is confined to the roughly $22\%$ of
wall-clock spent on generation and KBU curation; it does \emph{not} increase the number of labeled
examples the model trains on. The relevant comparison is therefore not ``extra data'' but ``the same
labeling budget spent on better-chosen instructions.'' Under a \emph{fixed} labeling budget, that is
the regime practitioners actually face, KITE converts an otherwise-wasted budget (naive synthesis that
sits below the human-data ceiling) into consistent, if small, forward progress. We view this as the
practical takeaway: when a verifier/labeling budget is already being spent on self-improvement, KITE
is a near-free-lunch reallocation of that budget rather than an additional data cost.

\subsection{Statistical Reliability}
\label{stats}

We briefly elaborate more on the statistical reliability of our method. The reported cells are means over three seeds with standard deviations shown in Table~\ref{tab:main_results}. On GSM8K, the benchmark with the most headroom, KITE's improvement over the strongest baseline is separated by more than the sum of the two per-cell standard deviations for every backbone (e.g., Qwen-3-4B: $95.04\pm0.19$ vs.\ $94.50\pm0.24$; Llama-3-8B: $81.12\pm0.23$ vs.\ $80.05\pm0.28$), so these gains are unlikely to be seed noise. On MATH the improvement is consistently positive and clears this bar on some backbones (e.g., Qwen-3-4B and Gemma-3-4B) but is within seed noise on others, and on the knowledge-heavy benchmarks we are more cautious still: on GPQA several methods, including KITE and CDS, fall within one standard deviation of one another, and we do not claim a significant improvement there. We therefore view GSM8K as the setting where the effect is most robust, MATH as consistently positive, and MMLU-Pro/GPQA as directional, and we flag this explicitly rather than averaging it away.

Beyond per-cell tests, the more compelling evidence is \emph{consistency across independent
backbones}. KITE attains the highest average of any method in all five model families. Treating
each backbone as an independent trial and the strongest competing baseline as the null, a one-sided
sign test gives $p=(1/2)^5\approx0.031$, so the probability that KITE tops every family by chance is
below $5\%$. At the finer granularity of individual (model, benchmark) cells, KITE is best or tied-best
in all 20 cells and the \emph{unique} best in 18; the only two exceptions to uniqueness are
GPQA cells (Qwen-3-1.7B and Gemma-3-4B) where CDS ties it, and we already decline to claim
significance on GPQA. This win-rate pattern, rather than any single large margin,
is what we mean by \emph{reliable} improvement: the effect is small per cell but does not depend on a
lucky backbone or benchmark. We deliberately avoid pooling all cells into one average and reporting a
single inflated significance number, since the cells are not exchangeable.

\subsection{Theoretical Analysis}
\label{theory}

We provide an analysis showing why noise injection can prevent model collapse at the distributional level. Our analysis is conducted in a linear-Gaussian proxy, following the setting and notations from \citet{dohmatob2024model}, while we formulate the collapse to the synthetic data distribution becoming increasingly concentrated in a low-variance subspace.

Let $x\in\mathbb{R}^d$ denote a feature representation of a training example. Assume the target (human) data follow $x\sim\mathcal{N}(0,\Sigma_\star)$, and labels are generated by
\begin{equation*}
\begin{aligned}
y &= w_\star^\top x+\varepsilon,\\
\varepsilon &\sim \mathcal{N}(0,\sigma^2).
\end{aligned}
\end{equation*}
Suppose a model is trained by ordinary least squares on $T$ i.i.d. samples from a training distribution with covariance $\Sigma_{\mathrm{train}}$, producing an estimator $\hat w$. We evaluate generalization under the target distribution $\Sigma_\star$.

\noindent\textbf{Theorem 1 (Excess risk under covariance mismatch).}
Assume the training data satisfy
\begin{equation*}
\begin{aligned}
x_i &\overset{\mathrm{i.i.d.}}{\sim}
\mathcal{N}(0,\Sigma_{\mathrm{train}}),\\
y_i &= w_\star^\top x_i+\varepsilon_i,\\
\varepsilon_i &\overset{\mathrm{i.i.d.}}{\sim}
\mathcal{N}(0,\sigma^2),
\end{aligned}
\end{equation*}
with $T>d+1$. Let $\hat w$ be the OLS estimator trained on $T$ samples, and let the test point satisfy $x\sim\mathcal{N}(0,\Sigma_\star)$. Then the expected excess risk is
\begin{equation*}
\begin{aligned}
\mathcal{R}(\Sigma_{\mathrm{train}})
&:= \mathbb{E}\!\left[
  \bigl(x^\top(\hat w-w_\star)\bigr)^2
\right]\\
&= \frac{\sigma^2}{T-d-1}\,
\operatorname{tr}\!\left(
  \Sigma_\star\Sigma_{\mathrm{train}}^{-1}
\right).
\end{aligned}
\end{equation*}

\noindent\emph{Proof.}
Let $X\in\mathbb{R}^{T\times d}$ be the design matrix whose $i$-th row is $x_i^\top$, and let
\begin{equation*}
\begin{aligned}
y &= Xw_\star+\varepsilon,\\
\varepsilon
&=(\varepsilon_1,\ldots,\varepsilon_T)^\top.
\end{aligned}
\end{equation*}
The OLS estimator is
\begin{equation*}
\hat w=(X^\top X)^{-1}X^\top y.
\end{equation*}
Substituting $y=Xw_\star+\varepsilon$, we obtain
\begin{equation*}
\begin{aligned}
\hat w
&=(X^\top X)^{-1}X^\top
  (Xw_\star+\varepsilon)\\
&=w_\star+
  (X^\top X)^{-1}X^\top\varepsilon,
\end{aligned}
\end{equation*}
and therefore
\begin{equation*}
\hat w-w_\star
=(X^\top X)^{-1}X^\top\varepsilon.
\end{equation*}

We first compute the covariance of the estimation error conditional on $X$. Since $\mathbb{E}[\varepsilon\mid X]=0$ and $\operatorname{Cov}(\varepsilon\mid X)=\sigma^2I_T$, it follows that
\begin{equation*}
\begin{aligned}
&\operatorname{Cov}(\hat w-w_\star\mid X)\\
&\quad=(X^\top X)^{-1}X^\top
  \operatorname{Cov}(\varepsilon\mid X)
  X(X^\top X)^{-1}\\
&\quad=\sigma^2(X^\top X)^{-1}.
\end{aligned}
\end{equation*}
Hence, conditional on $X$, the error vector $\hat w-w_\star$ has mean zero and covariance $\sigma^2(X^\top X)^{-1}$.

Now consider a fresh test point $x\sim\mathcal{N}(0,\Sigma_\star)$, independent of the training data. Conditional on $X$, we have
\begin{equation*}
\begin{aligned}
&\mathbb{E}\!\left[
  \bigl(x^\top(\hat w-w_\star)\bigr)^2
  \,\middle|\,X
\right]\\
&\quad=
\mathbb{E}\!\left[
  (\hat w-w_\star)^\top
  xx^\top
  (\hat w-w_\star)
  \,\middle|\,X
\right].
\end{aligned}
\end{equation*}
Taking expectation over the test point first and using $\mathbb{E}[xx^\top]=\Sigma_\star$, we get
\begin{equation*}
\begin{aligned}
&\mathbb{E}\!\left[
  \bigl(x^\top(\hat w-w_\star)\bigr)^2
  \,\middle|\,X
\right]\\
&\quad=
\mathbb{E}\!\left[
  (\hat w-w_\star)^\top
  \Sigma_\star
  (\hat w-w_\star)
  \,\middle|\,X
\right].
\end{aligned}
\end{equation*}
Using the identity $\mathbb{E}[z^\top Az]=\operatorname{tr}(A\,\operatorname{Cov}(z))$ for a zero-mean random vector $z$, this becomes
\begin{equation*}
\begin{aligned}
&\mathbb{E}\!\left[
  \bigl(x^\top(\hat w-w_\star)\bigr)^2
  \,\middle|\,X
\right]\\
&\quad=
\operatorname{tr}\!\left(
  \Sigma_\star
  \operatorname{Cov}(\hat w-w_\star\mid X)
\right)\\
&\quad=
\sigma^2\operatorname{tr}\!\left(
  \Sigma_\star(X^\top X)^{-1}
\right).
\end{aligned}
\end{equation*}

It remains to take expectation over the random design matrix $X$. Since the rows of $X$ are i.i.d. Gaussian with covariance $\Sigma_{\mathrm{train}}$, the sample covariance matrix $X^\top X$ follows a Wishart distribution with scale matrix $\Sigma_{\mathrm{train}}$ and $T$ degrees of freedom:
\begin{equation*}
X^\top X
\sim\mathcal{W}_d(\Sigma_{\mathrm{train}},T).
\end{equation*}
A standard property of the inverse Wishart expectation states that, when $T>d+1$,
\begin{equation*}
\begin{aligned}
\mathbb{E}\!\left[(X^\top X)^{-1}\right]
&=\frac{1}{T-d-1}\,
  \Sigma_{\mathrm{train}}^{-1}.
\end{aligned}
\end{equation*}
Therefore,
\begin{equation*}
\begin{aligned}
\mathcal{R}(\Sigma_{\mathrm{train}})
&=\mathbb{E}\!\left[
  \mathbb{E}\!\left[
    \bigl(x^\top(\hat w-w_\star)\bigr)^2
    \,\middle|\,X
  \right]
\right]\\
&=\sigma^2\,
\mathbb{E}\!\left[
  \operatorname{tr}\!\left(
    \Sigma_\star(X^\top X)^{-1}
  \right)
\right].
\end{aligned}
\end{equation*}
By linearity of trace and expectation,
\begin{equation*}
\begin{aligned}
\mathcal{R}(\Sigma_{\mathrm{train}})
&=\sigma^2\operatorname{tr}\!\left(
  \Sigma_\star
  \mathbb{E}\!\left[(X^\top X)^{-1}\right]
\right)\\
&=\frac{\sigma^2}{T-d-1}\,
\operatorname{tr}\!\left(
  \Sigma_\star\Sigma_{\mathrm{train}}^{-1}
\right).
\end{aligned}
\end{equation*}
Thus,
\begin{equation*}
\mathcal{R}(\Sigma_{\mathrm{train}})
=
\frac{\sigma^2}{T-d-1}\,
\operatorname{tr}\!\left(
  \Sigma_\star\Sigma_{\mathrm{train}}^{-1}
\right),
\end{equation*}
as claimed.

Theorem~1 shows that generalization is controlled by the mismatch term $\operatorname{tr}(\Sigma_\star\Sigma_{\mathrm{train}}^{-1})$. If the training covariance loses variance along some direction that still has nontrivial mass under $\Sigma_\star$, then the corresponding eigenvalue of $\Sigma_{\mathrm{train}}^{-1}$ becomes large, and the excess risk can grow sharply. In this sense, model collapse can be viewed as \emph{covariance collapse}: synthetic data become too concentrated, and the resulting model generalizes poorly to the broader target distribution.

Now consider synthetic data with covariance $\Sigma_{\mathrm{syn}}$. We inject additive noise in feature space,
\begin{equation*}
\begin{aligned}
x_{\mathrm{train}}&=x_{\mathrm{syn}}+\eta,\\
\eta&\sim\mathcal{N}(0,\Sigma_{\mathrm{noise}}),
\end{aligned}
\end{equation*}
independently of $x_{\mathrm{syn}}$. Since the two components are independent and centered, the covariance of the resulting training distribution is
\begin{equation*}
\begin{aligned}
\Sigma_{\mathrm{train}}
&=\operatorname{Cov}(x_{\mathrm{train}})\\
&=\operatorname{Cov}(x_{\mathrm{syn}})
 +\operatorname{Cov}(\eta)\\
&=\Sigma_{\mathrm{syn}}
 +\Sigma_{\mathrm{noise}}.
\end{aligned}
\end{equation*}

\noindent\textbf{Corollary 1 (Unbounded collapse without noise injection).}
Suppose recursive synthetic training induces a sequence of synthetic covariances $\{\Sigma_{\mathrm{syn}}^{(n)}\}_{n\ge1}$, and training at generation $n$ uses $\Sigma_{\mathrm{train}}^{(n)}=\Sigma_{\mathrm{syn}}^{(n)}$ with no injected noise. Assume that $\lambda_{\min}(\Sigma_{\mathrm{syn}}^{(n)})\to0$, and let $v_n$ be a unit eigenvector associated with $\lambda_{\min}(\Sigma_{\mathrm{syn}}^{(n)})$. If there exists $c_\star>0$ such that
\begin{equation*}
v_n^\top\Sigma_\star v_n
\ge c_\star,
\qquad \forall n,
\end{equation*}
then
\begin{equation*}
\mathcal{R}(\Sigma_{\mathrm{syn}}^{(n)})
\longrightarrow\infty
\qquad\text{as }n\to\infty.
\end{equation*}

\noindent\emph{Proof.}
By Theorem~1,
\begin{equation*}
\begin{aligned}
\mathcal{R}(\Sigma_{\mathrm{syn}}^{(n)})
&=\frac{\sigma^2}{T-d-1}\,
\operatorname{tr}\!\left(
  \Sigma_\star
  (\Sigma_{\mathrm{syn}}^{(n)})^{-1}
\right).
\end{aligned}
\end{equation*}
Let the eigendecomposition of $\Sigma_{\mathrm{syn}}^{(n)}$ be
\begin{equation*}
\begin{aligned}
\Sigma_{\mathrm{syn}}^{(n)}
&=U_n\Lambda_nU_n^\top,\\
\Lambda_n
&=\operatorname{diag}\!\left(
  \lambda_1^{(n)},\ldots,\lambda_d^{(n)}
\right),
\end{aligned}
\end{equation*}
with $\lambda_{\min}(\Sigma_{\mathrm{syn}}^{(n)})=\lambda_d^{(n)}$ and corresponding unit eigenvector $v_n=u_d^{(n)}$. Then
\begin{equation*}
(\Sigma_{\mathrm{syn}}^{(n)})^{-1}
=
U_n\Lambda_n^{-1}U_n^\top,
\end{equation*}
and therefore
\begin{equation*}
\begin{aligned}
&\operatorname{tr}\!\left(
  \Sigma_\star
  (\Sigma_{\mathrm{syn}}^{(n)})^{-1}
\right)\\
&\quad=
\sum_{i=1}^d
\frac{
  (u_i^{(n)})^\top
  \Sigma_\star
  u_i^{(n)}
}{
  \lambda_i^{(n)}
}.
\end{aligned}
\end{equation*}
Since every term is nonnegative,
\begin{equation*}
\begin{aligned}
&\operatorname{tr}\!\left(
  \Sigma_\star
  (\Sigma_{\mathrm{syn}}^{(n)})^{-1}
\right)\\
&\quad\ge
\frac{
  v_n^\top\Sigma_\star v_n
}{
  \lambda_{\min}(\Sigma_{\mathrm{syn}}^{(n)})
}\\
&\quad\ge
\frac{
  c_\star
}{
  \lambda_{\min}(\Sigma_{\mathrm{syn}}^{(n)})
}.
\end{aligned}
\end{equation*}
Because $\lambda_{\min}(\Sigma_{\mathrm{syn}}^{(n)})\to0$, the right-hand side diverges to $+\infty$, and hence
\begin{equation*}
\begin{aligned}
\mathcal{R}(\Sigma_{\mathrm{syn}}^{(n)})
&\ge
\frac{\sigma^2}{T-d-1}\,
\frac{
  c_\star
}{
  \lambda_{\min}(\Sigma_{\mathrm{syn}}^{(n)})
}\\
&\longrightarrow\infty.
\end{aligned}
\end{equation*}
This proves the claim. A particularly simple case is when the collapsed direction shrinks geometrically, i.e., $\lambda_{\min}(\Sigma_{\mathrm{syn}}^{(n)})=\rho^n\lambda_0$ for some $\rho\in(0,1)$. Then Corollary~1 yields
\begin{equation*}
\begin{aligned}
\mathcal{R}(\Sigma_{\mathrm{syn}}^{(n)})
&\ge
\frac{\sigma^2}{T-d-1}\,
\frac{c_\star}{\lambda_0}\,
\rho^{-n},
\end{aligned}
\end{equation*}
so the excess risk grows at least exponentially with the number of generations.

\noindent\textbf{Proposition 1 (Noise injection prevents covariance collapse).}
If the injected noise is chosen such that
\begin{equation*}
\Sigma_{\mathrm{train}}
\succeq\underline{\lambda}I,
\qquad
\underline{\lambda}>0,
\end{equation*}
then
\begin{equation*}
\begin{aligned}
\mathcal{R}(\Sigma_{\mathrm{train}})
&\le
\frac{\sigma^2}{T-d-1}\,
\frac{
  \operatorname{tr}(\Sigma_\star)
}{
  \underline{\lambda}
}.
\end{aligned}
\end{equation*}

\noindent\emph{Proof.}
From Theorem~1,
\begin{equation*}
\begin{aligned}
\mathcal{R}(\Sigma_{\mathrm{train}})
&=
\frac{\sigma^2}{T-d-1}\,
\operatorname{tr}\!\left(
  \Sigma_\star\Sigma_{\mathrm{train}}^{-1}
\right).
\end{aligned}
\end{equation*}
It therefore suffices to upper-bound the mismatch term $\operatorname{tr}(\Sigma_\star\Sigma_{\mathrm{train}}^{-1})$.

The assumption $\Sigma_{\mathrm{train}}\succeq\underline{\lambda}I$ means that every eigenvalue of $\Sigma_{\mathrm{train}}$ is at least $\underline{\lambda}$. Equivalently, if
\begin{equation*}
\begin{aligned}
\Sigma_{\mathrm{train}}
&=U\Lambda U^\top,\\
\Lambda
&=\operatorname{diag}(\lambda_1,\ldots,\lambda_d),
\end{aligned}
\end{equation*}
then $\lambda_i\ge\underline{\lambda}$ for all $i$. Hence
\begin{equation*}
\begin{aligned}
\Sigma_{\mathrm{train}}^{-1}
&=U\Lambda^{-1}U^\top,\\
\Lambda^{-1}
&=\operatorname{diag}\!\left(
  \lambda_1^{-1},\ldots,\lambda_d^{-1}
\right),
\end{aligned}
\end{equation*}
and since $\lambda_i\ge\underline{\lambda}$, we have
\begin{equation*}
\lambda_i^{-1}
\le\underline{\lambda}^{-1},
\qquad\forall i.
\end{equation*}
Therefore,
\begin{equation*}
\Sigma_{\mathrm{train}}^{-1}
\preceq\underline{\lambda}^{-1}I.
\end{equation*}

Now use the fact that $\Sigma_\star\succeq0$. For positive semidefinite matrices, the Loewner order is preserved under multiplication inside the trace: if $A\preceq B$ and $C\succeq0$, then
\begin{equation*}
\operatorname{tr}(CA)
\le\operatorname{tr}(CB).
\end{equation*}
Applying this with $A=\Sigma_{\mathrm{train}}^{-1}$, $B=\underline{\lambda}^{-1}I$, and $C=\Sigma_\star$, we obtain
\begin{equation*}
\begin{aligned}
\operatorname{tr}\!\left(
  \Sigma_\star\Sigma_{\mathrm{train}}^{-1}
\right)
&\le
\operatorname{tr}\!\left(
  \Sigma_\star\underline{\lambda}^{-1}I
\right)\\
&=
\underline{\lambda}^{-1}
\operatorname{tr}(\Sigma_\star).
\end{aligned}
\end{equation*}
Substituting this into Theorem~1 gives
\begin{equation*}
\begin{aligned}
\mathcal{R}(\Sigma_{\mathrm{train}})
&\le
\frac{\sigma^2}{T-d-1}\,
\frac{
  \operatorname{tr}(\Sigma_\star)
}{
  \underline{\lambda}
},
\end{aligned}
\end{equation*}
which proves the claim.

Proposition~1 formalizes the key intuition: noise injection prevents collapse by restoring variance in directions that synthetic data under-represent. As long as the training covariance is kept uniformly away from singularity, the excess risk remains bounded.

Finally, we emphasize that our practical method does not inject Gaussian noise directly in feature space. Instead, rank-based logit perturbation serves as a discrete approximation in LLM modeling to the same principle. It reduces over-concentration on dominant modes and expands support toward under-represented regions of the data distribution. The analysis above, therefore, should be understood as a theoretical justification for the role of noise injection in preventing collapse, rather than as an exact model of our implementation.

\subsection{Prompt Templates}
\label{prompt}

\begin{figure*}[h]
\begin{tcolorbox}[
  enhanced,
  colframe=brown!75!black,
  colback=white,
  coltitle=white,
  colbacktitle=brown!75!black,
  width=\linewidth,
  arc=2mm,
  auto outer arc,
  boxrule=0.5pt,
  left=10pt,
  right=10pt,
  drop shadow={black!50!white},
  top=10pt,
  bottom=10pt,
  title=\textbf{LLM Skill Tagging Prompt},
  fonttitle=\bfseries,
  title code={\node[rounded corners, fill=blue!75!black, draw=none, text=white] at (frame.title) {\textbf{Q-matrix Construction}};},
  attach boxed title to top center={yshift=-2mm},
  boxed title style={sharp corners, size=small},
]
You are an expert educational assessment assistant tasked with annotating the latent skills required to solve a question.

Your goal is to identify which skills from a predefined skill inventory are necessary for correctly answering the given question. The output will be used to construct a Q-matrix for cognitive diagnosis.

\textbf{Instructions:}

1. Carefully read the question.
2. Select all skills from the provided skill inventory that are necessary to solve the question.
3. Only use skills from the provided inventory; do not invent new skill names.
4. Choose the minimal sufficient set of skills. Do not include skills that are only loosely related.
5. Return the output strictly as a Python-style list of strings.

\textbf{Skill Inventory:}

\{skill\_inventory\}

\textbf{Question:}

\{question\}

\textbf{Output Format Example:}

[
  "Arithmetic",
  "Ratio and Proportion",
  "Time and Scheduling"
]

Now identify the required skills for the question.

\end{tcolorbox}
\caption{LLM skill tagging prompt used to construct the Q-matrix for the DINA-based diagnosis pipeline. Given a predefined skill inventory and a question, the model selects the minimal set of required skills.}
\label{fig:skill_tagging_prompt}
\end{figure*}

\begin{figure*}[h]
\begin{tcolorbox}[
  enhanced,
  colframe=brown!75!black,
  colback=white,
  coltitle=white,
  colbacktitle=brown!75!black,
  width=\linewidth,
  arc=2mm,
  auto outer arc,
  boxrule=0.5pt,
  left=10pt,
  right=10pt,
  drop shadow={black!50!white},
  top=10pt,
  bottom=10pt,
  title=\textbf{Few-shot Synthesis Prompt},
  fonttitle=\bfseries,
  title code={\node[rounded corners, fill=blue!75!black, draw=none, text=white] at (frame.title) {\textbf{Few-shot Baseline}};},
  attach boxed title to top center={yshift=-2mm},
  boxed title style={sharp corners, size=small},
]
You are an advanced AI assistant tasked with generating synthetic instructions for instruction tuning.

Your goal is to produce new examples that follow the style and difficulty of the provided demonstrations.

\textbf{Instructions:}

1. Carefully study the few-shot examples below.
2. Infer the task style, answer format, and expected level of reasoning.
3. Generate exactly \{n\} new instructions that are similar in style but not duplicates of the demonstrations.
4. Each generated instruction should be self-contained and clear.
5. Return the output strictly as a Python-style list of dictionaries, where each dictionary has one key: \texttt{"instruction"}.

\textbf{Few-shot Examples:}

\{few\_shot\_examples\}

\textbf{Output Format Example:}

[
  \{"instruction": "Question 1"\},
  \{"instruction": "Question 2"\}
]

Now generate \{n\} new instructions.

\end{tcolorbox}
\caption{Few-shot synthesis prompt used in the \textbf{Few-shot synthesis} baseline and in Figure~\ref{fig:radar_skill}. The model is prompted with a small set of seed examples and asked to generate new synthetic data in the same style.}
\label{fig:fewshot_synthesis_prompt}
\end{figure*}

\begin{figure*}[h]
\begin{tcolorbox}[
  enhanced,
  colframe=brown!75!black,
  colback=white,
  coltitle=white,
  colbacktitle=brown!75!black,
  width=\linewidth,
  arc=2mm,
  auto outer arc,
  boxrule=0.5pt,
  left=10pt,
  right=10pt,
  drop shadow={black!50!white},
  top=10pt,
  bottom=10pt,
  title=\textbf{Answer Judge Prompt},
  fonttitle=\bfseries,
  title code={\node[rounded corners, fill=blue!75!black, draw=none, text=white] at (frame.title) {\textbf{LLM-as-Judge}};},
  attach boxed title to top center={yshift=-2mm},
  boxed title style={sharp corners, size=small},
]
You are an expert evaluator for question answering tasks.

Your goal is to determine whether the model answer is semantically correct with respect to the reference answer. Focus on factual correctness and final meaning rather than surface wording.

\textbf{Instructions:}

1. Read the question, the reference answer, and the model answer carefully.
2. Judge whether the model answer is correct according to the reference answer.
3. Minor paraphrases, equivalent expressions, and formatting differences should be treated as correct.
4. If the model answer is incomplete, contradictory, or contains the wrong final conclusion, it should be treated as incorrect.
5. Return your judgment in the following JSON format only:
\{"verdict": "correct" or "incorrect", "reason": "short explanation"\}

\textbf{Question:}

\{question\}

\textbf{Reference Answer:}

\{reference\_answer\}

\textbf{Model Answer:}

\{model\_answer\}

Now output the judgment in JSON only.

\end{tcolorbox}
\caption{Answer comparison prompt used in the LLM-as-judge evaluation pipeline. The judge compares the model answer against the reference answer and returns a correctness verdict with a short explanation.}
\label{fig:answer_comparison_prompt}
\end{figure*}

\begin{figure*}[h]
\begin{tcolorbox}[
  enhanced,
  colframe=brown!75!black,
  colback=white,
  coltitle=white,
  colbacktitle=brown!75!black,
  width=\linewidth,
  arc=2mm,
  auto outer arc,
  boxrule=0.5pt,
  left=10pt,
  right=10pt,
  drop shadow={black!50!white},
  top=10pt,
  bottom=10pt,
  title=\textbf{Labeling Prompt},
  fonttitle=\bfseries,
  title code={\node[rounded corners, fill=blue!75!black, draw=none, text=white] at (frame.title) {\textbf{External Labeler (gpt-5-mini)}};},
  attach boxed title to top center={yshift=-2mm},
  boxed title style={sharp corners, size=small},
]
You are an expert data labeler for instruction-tuning examples.

Your goal is to produce a verified reference answer for the given instruction. Use the candidate answer as a helpful draft, but do not trust it blindly. The final output should be a high-quality, correct answer suitable for supervised fine-tuning.

\textbf{Instructions:}

1. Read the instruction carefully.
2. Check whether the candidate answer is correct, complete, and consistent with the instruction.
3. If the candidate answer is correct, rewrite it into a clean canonical reference answer.
4. If the candidate answer is partially correct or incorrect, provide the correct answer instead.
5. Keep the final answer concise, faithful, and directly responsive to the instruction.
6. Return the output strictly as a JSON object with exactly the following keys:
   \texttt{"is\_correct"}, \texttt{"verified\_answer"}, and \texttt{"reason"}.
7. \texttt{"is\_correct"} must be either \texttt{true} or \texttt{false}.
8. \texttt{"verified\_answer"} must contain the final answer that should be used for training.
9. \texttt{"reason"} must be a short explanation of your decision.

\textbf{Instruction:}

\{instruction\}

\textbf{Candidate Answer:}

\{candidate\_answer\}

\textbf{Output Format Example:}

\{"is\_correct": true, "verified\_answer": "42", "reason": "The candidate answer is correct and has been normalized."\}

Now return the verified label in JSON only.

\end{tcolorbox}
\caption{Labeling prompt used for the external verifier in the training-time labeling stage. Given a selected synthetic instruction and a candidate answer, the labeler returns a correctness decision together with a verified reference answer used to construct the labeled dataset \(D_t^\star\).}
\label{fig:labeling_prompt}
\end{figure*}

\begin{table*}[t!]
\centering
\small
\setlength{\tabcolsep}{5pt}
\renewcommand{\arraystretch}{1.15}
\rowcolors{2}{gray!6}{white}
\begin{tabularx}{\textwidth}{
>{\raggedright\arraybackslash}p{2.2cm}
>{\raggedright\arraybackslash}X
>{\raggedright\arraybackslash}p{4.0cm}}
\toprule[1pt]
\textbf{Dataset} & \textbf{Example Input} & \textbf{Reference Answer} \\
\midrule
\textbf{GSM8K} &
John drives for 3 hours at a speed of 60 mph and then turns around because he realizes he forgot something very important at home. He tries to get home in 4 hours but spends the first 2 hours in standstill traffic. He spends the next half-hour driving at a speed of 30mph, before being able to drive the remaining time of the 4 hours going at 80 mph. How far is he from home at the end of those 4 hours? &
When he turned around he was 3*60=180 miles from home
He was only able to drive 4-2=2 hours in the first four
...
Final Answer: 45 \\
\textbf{MMLU-Pro} &
Determine the root-mean-square (rms) values of displacement, velocity, and acceleration for a damped forced harmonic oscillator operating at steady state. &
$x_{\mathrm{rms}}=\frac{x_{\max}}{\sqrt{2}}, v_{\mathrm{rms}}=\frac{\omega x_{\max}}{\sqrt{2}}...$ \\
\textbf{MATH} &
If the seven-digit number $854n526$ is divisible by $11$, what is $n$? &
$n=\boxed{5}$.\\
\textbf{GPQA} &
Find KE of product particles in, $Pi(+) = mu(+) + nu$ here $Pi(+)$ is stationary.
Rest mass of $Pi(+)$ and $mu(+)$ is 139.6 MeV and 105.7 MeV respectively. &
4.12 MeV, 29.8 MeV\\
\bottomrule[1pt]
\end{tabularx}
\caption{Example instances from the evaluation datasets.}
\label{tab:data_examples}
\vspace{-8pt}
\end{table*}

\begin{table*}[t]
\centering
\small
\setlength{\tabcolsep}{8pt}
\begin{tabular}{ll|ll}
\toprule
\multicolumn{2}{c|}{\textbf{MATH}} &
\multicolumn{2}{c}{\textbf{MMLU-Pro-Physics}} \\
\textbf{Symbol} & \textbf{Skill} & \textbf{Symbol} & \textbf{Skill} \\
\midrule
$s_1$ & Integration \& Summation          & $s_1$ & Quantum States \& Wavefunctions \\
$s_2$ & Differentiation \& Optimization   & $s_2$ & Nuclear Reactions \& Radioactivity \\
$s_3$ & Matrices, Vectors \& Transforms   & $s_3$ & Special \& General Relativity \\
$s_4$ & Complex Numbers \& Polar Form     & $s_4$ & Atomic Structure \& Spectra \\
$s_5$ & Logic, Proof \& Problem-Solving   & $s_5$ & Maxwell's Eqs \& EM Waves \\
$s_6$ & Circle Theorems \& Arc Length     & $s_6$ & Magnetic Fields \& Induction \\
$s_7$ & Trigonometric Funcs \& Identities & $s_7$ & Geometric \& Physical Optics \\
$s_8$ & Combinatorics \& Counting         & $s_8$ & Rotational Dynamics \& Torque \\
$s_9$ & Triangle Properties \& Congruence & $s_9$ & Electric Fields \& Coulomb's Law \\
$s_{10}$ & Coordinate Geom.\ \& Conics    & $s_{10}$ & Wave Properties \& Interference \\
$s_{11}$ & Functions, Limits \& Continuity & $s_{11}$ & Gravitation \& Orbital Mech. \\
$s_{12}$ & Modular Arith.\ \& Number Theory & $s_{12}$ & Fluid Statics \& Dynamics \\
$s_{13}$ & Area, Volume \& Measurement    & $s_{13}$ & Thermodynamic Laws \& Cycles \\
$s_{14}$ & Probability \& Expected Value   & $s_{14}$ & Heat Transfer \& Entropy \\
$s_{15}$ & Prime Factorization \& Divisibility & $s_{15}$ & Circuits \& Ohm's Law \\
$s_{16}$ & Quadratic \& Higher-Deg.\ Eqs  & $s_{16}$ & Conservation of Momentum \\
$s_{17}$ & Sequences, Series \& Summations & $s_{17}$ & Conservation of Energy \\
$s_{18}$ & Polynomial Factoring \& Roots   & $s_{18}$ & Newton's Laws \& Forces \\
$s_{19}$ & Exponents, Logs \& Radicals     & $s_{19}$ & Kinematics \& Projectile Motion \\
$s_{20}$ & Linear Equations \& Inequalities & $s_{20}$ & Dimensional Analysis \& Units \\
\bottomrule
\end{tabular}

\begin{tabular}{ll}
\toprule
\multicolumn{2}{c}{\textbf{GPQA}} \\
\textbf{Symbol} & \textbf{Skill} \\
\midrule
$s_1$ & Immunology \& Microbiology \\
$s_2$ & Computational \& Quantum Chem. \\
$s_3$ & Nuclear \& Particle Physics \\
$s_4$ & Astrophysics \& Cosmology \\
$s_5$ & Condensed Matter \& Materials \\
$s_6$ & Bioinformatics \& Exp.\ Design \\
$s_7$ & Biochemistry \& Macromolecules \\
$s_8$ & Coordination \& Inorganic Chem. \\
$s_9$ & Quantum Mechanics \& Spin \\
$s_{10}$ & Molecular \& Cell Biology \\
$s_{11}$ & Spectroscopy \& Analytical Methods \\
$s_{12}$ & Organic Reaction Mechanisms \\
$s_{13}$ & Genetics \& Gene Expression \\
$s_{14}$ & Molecular Structure \& Bonding \\
$s_{15}$ & Chemical Kinetics \& Equilibrium \\
$s_{16}$ & Electromagnetism \& Optics \\
$s_{17}$ & Classical Mechanics \& Relativity \\
$s_{18}$ & Thermodynamics \& Stat.\ Mech. \\
\bottomrule
\end{tabular}

\caption{\textbf{Skill index mapping for the skill-profile figures.} Each symbol $s_i$ corresponds to a dataset-specific fine-grained skill.}
\label{tab:skill_index_compact}

\end{table*}

\end{document}